\documentclass[Afour, sageh, times]{sagej}
\usepackage{times}
\usepackage{moreverb,url}
\usepackage{graphicx}%
\usepackage{multirow}%
\usepackage{amsmath,amssymb,amsfonts}%
\usepackage{amsthm}%
\usepackage{mathrsfs}%
\usepackage[title]{appendix}%
\usepackage{xcolor}%
\usepackage{textcomp}%
\usepackage{rotating}
\usepackage{manyfoot}%
\usepackage{pifont}
\usepackage{booktabs}%
\usepackage{algorithm}%
\usepackage{algorithmicx}%
\usepackage{subcaption}%
\usepackage{adjustbox}  
\usepackage{listings}%
\usepackage{titlesec}
\usepackage{array}

\titleformat{\subsubsection}[block]{\normalsize\bfseries}{\thesubsubsection}{1em}{}

\usepackage[colorlinks,bookmarksopen,bookmarksnumbered,citecolor=red,urlcolor=red]{hyperref}

\usepackage[font=small]{caption}

\def\secref#1{Sec.~\ref{#1}}
\def\figref#1{Fig.~\ref{#1}}
\def\tabref#1{Tab.~\ref{#1}}
\def\eqref#1{Eq.~(\ref{#1})}

\makeatletter
\usepackage{xspace}
\DeclareRobustCommand\onedot{\futurelet\@let@token\@onedot}
\def\@onedot{\ifx\@let@token.\else.\null\fi\xspace}

\makeatother

\newcolumntype{L}[1]{>{\raggedright\let\newline\\\arraybackslash\hspace{0pt}}m{#1}}
\newcolumntype{C}[1]{>{\centering\let\newline\\\arraybackslash\hspace{0pt}}m{#1}}
\newcolumntype{R}[1]{>{\raggedleft\let\newline\\\arraybackslash\hspace{0pt}}m{#1}}

\def\argmax{\mathop{\rm argmax}}

\newcommand{\norm}[1]{\lVert#1\lVert}

\renewcommand{\b}[1]{\mbox{\boldmath$#1$}}

\newcommand{\logit}{\b{\ell}}
\newcommand{\cmark}{\ding{51}}%
\newcommand{\xmark}{\ding{55}}%
\begin{document}
\runninghead{Sodano et al.}

\title{\LARGE \bf Open-World Panoptic Segmentation}

\author{Matteo Sodano\affilnum{1} and Federico Magistri\affilnum{1} and Jens Behley\affilnum{1} and Cyrill Stachniss\affilnum{1,2}}

\affiliation{\affilnum{1}Center for Robotics, University of Bonn,  Germany\\
\affilnum{2}Lamarr Institute for Machine Learning and Artificial Intelligence, Germany
}

\corrauth{Matteo Sodano,
Center for Robotics,
University of Bonn,
Germany.}

\email{matteo.sodano7@gmail.com}

\begin{abstract}
	Robots need to be able to understand their surroundings in order to operate safely and robustly, and to interact with the surrounding environment. Robots deployed in unconstrained real-world scenarios must additionally be able to deal with novel situations and objects that have never been seen before. In this article, we tackle the problem of open-world panoptic segmentation, i.e., the task of discovering new semantic categories and new object instances at test time, while enforcing consistency among the categories that we incrementally discover.
	We present Con2MAV, a general method for open-world panoptic segmentation. Experiments across a wide range of datasets, from road scenes to underwater environments, highlight its compelling capabilities in open-world segmentation and its competitive performance on known classes. We will open-source the implementation of our approach upon acceptance. In addition, we propose PANIC (\textbf{P}anoptic \textbf{AN}omalies \textbf{I}n \textbf{C}ontext), a benchmark for evaluating open-world segmentation tasks in autonomous driving scenarios. This dataset, recorded with a multi-modal sensor suite mounted on a car, and then manually annotated, provides high-quality, pixel-wise annotations of anomalous objects at both semantic and instance level. PANIC contains 800 images, more than 50 unknown classes, i.e., classes that do not appear in the training set, and over 4,000 object instances, providing a comprehensive benchmark for evaluating open-world segmentation methods in autonomous driving scenarios. We provide competitions for multiple open-world segmentation tasks on a hidden test set. Our dataset and competitions are available at \url{https://www.ipb.uni-bonn.de/data/panic}.	\vspace{-0.25cm}
\end{abstract}


\maketitle

\section{Introduction}\label{sec:intro}

Interpreting sensor data is a key capability needed by robots operating in unconstrained real-world scenarios. Robots, in fact, must be able to understand their surroundings to operate safely and robustly. For this reason, we witnessed enormous progress in scene interpretation tasks such as object detection~\citep{girshick2015iccv, ren2016tpami}, semantic segmentation~\citep{milioto2019icra, oliveira2018ijrr}, instance segmentation~\citep{he2017iccv, marks2023ral}, and panoptic segmentation~\citep{cheng2022cvpr, kirillov2019cvpr, roggiolani2023icra, sodano2023icra}. However, these tasks in their conventional setting share the limiting assumption that all classes and objects that appear at inference time have already been seen at training time. This so-called \textit{closed-world} assumption greatly limits the deployment of such algorithms in the real world. A central challenge for learning-based systems is the ability to deal with something that they have never seen before at training time. For example, autonomous cars navigating in the real world will eventually experience situations and objects that they have not seen before, and need to be able to identify them in order to operate safely and not cause any harm. Thus, autonomous vision systems should be able to deal with the \textit{open-world} scenario, i.e., the fact that not everything can be covered in the training data and novel objects and categories should be identified at test time. 

\begin{figure*}[t!]
    \centering
      \includegraphics[width=\textwidth]{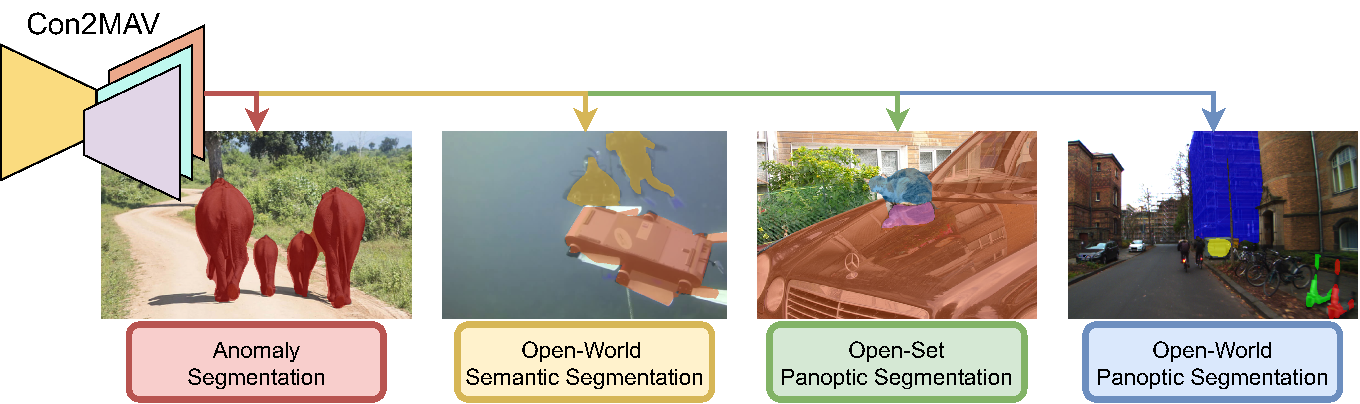}
      \caption{Our proposed approach, Con2MAV, is able to tackle multiple open-world tasks and segment unknown objects and categories. In the figure, we show predictions on (left to right) SegmentMeIfYouCan~\citep{chan2021neurips}, SUIM~\citep{islam2020iros}, COCO~\citep{lin2014eccv}, and PANIC (ours). \vspace{-0.5em}}
      \label{fig:model_motivation}
\end{figure*}

Recently, deep learning-based methods have achieved remarkable performance in semantic scene understanding~\citep{cheng2020cvpr, cheng2022cvpr, tang2021cvpr}. However, they typically operate under a closed-world assumption, predicting only within a fixed set of categories defined during training. As a result, they cannot recognize objects that fall outside this predefined taxonomy and often misclassify unknown instances with high confidence.
Open-vocabulary, also called zero-shot segmentation, approaches~\citep{carion2025arxiv, ghiasi2022eccv} attempt to relax this constraint by leveraging text embeddings from large-scale vision-language models. However, they still require all relevant categories to be anticipated and explicitly included in the vocabulary via prompts before segmenting an image. This fundamentally limits their applicability in dynamic, unpredictable environments where novel objects may appear without any prior semantic description.

Open-world panoptic segmentation enables a level of scene understanding that goes beyond both standard panoptic segmentation and open-vocabulary approaches. Unlike closed-world panoptic segmentation, which is restricted to a fixed set of predefined classes, the open-world setting requires the model to discover and segment previously unseen categories and their individual object instances at test time. And unlike open-vocabulary panoptic segmentation, which can only recognize categories that are explicitly provided through text prompts or belong to the vocabulary of a vision-language model, open-world panoptic segmentation makes no assumptions about which and how many classes may appear. This capability is essential for real-world autonomous systems, which must operate robustly in unpredictable, dynamic, and open environments where new objects or categories emerge at any time.

Thus, in this article, we address open-world panoptic segmentation. We introduce Con2MAV, a fully convolutional method that leverages a new loss function to construct robust pre-logit class descriptors, enabling meaningful representations even when only few training categories are available. Beyond segmenting unknown objects, Con2MAV incrementally discovers novel semantic categories at test time and maintains consistent category identities across images through an online descriptor update mechanism. Our formulation is architecture-agnostic and can be applied to any backbone. Con2MAV achieves compelling performance on anomaly segmentation, open-world semantic segmentation, and open-world panoptic segmentation across several benchmarks, including SegmentMeIfYouCan~\citep{chan2021neurips}, COCO~\citep{lin2014eccv}, BDDAnomaly~\citep{hendrycks2022icml}, and SUIM~\citep{islam2020iros}. Illustrative predictions are shown in~\figref{fig:model_motivation}.

\begin{figure*}[t!]
    \centering
    \includegraphics[width=0.95\textwidth]{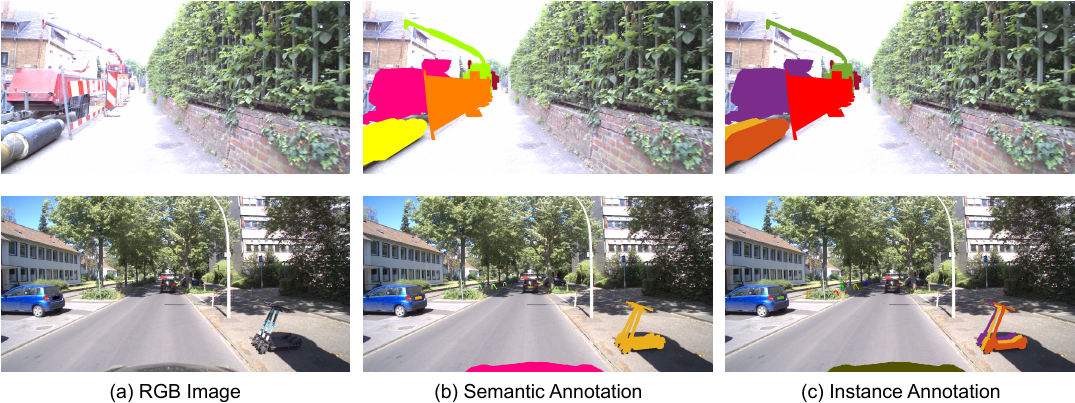}
    \caption{Our dataset, PANIC, provides pixel-wise annotations of unknown semantic categories and object instances of RGB images. The images have been recorded with our own inhouse sensor suite~\citep{vizzo2023itsc} mounted on our vehicle driving in Bonn, Germany. The dataset consists of images collected at different times of day over the span of more than a year. \vspace{-1em}}
    \label{fig:data_motivation}
  \end{figure*}

Additionally, motivated by recent advances in open-world segmentation~\citep{delic2024arxiv, grcic2023cvpr, nayal2023iccv, rai2024tpami, sodano2024cvpr}, we introduce the PANIC (Panoptic Anomalies In Context) benchmark. Recorded in Bonn, Germany, with our sensor suite~\citep{vizzo2023itsc} mounted on a car, PANIC includes 800 images with semantic and instance labels for all anomalous objects. We consider objects that do not appear in Cityscapes~\citep{cordts2016cvpr} as anomalies. They can appear anywhere in the image and have any size. We split the dataset in validation and test images, and no unknown class that appears in the test set is present in the validation set, making the test set truly hidden. Examples from the dataset are shown in~\figref{fig:data_motivation}.

Lastly, we provide an extended discussion on open-world segmentation tasks. Beside anomaly segmentation, which is quite common in the literature, other open-world tasks suffer from an ambiguous task nomenclature, and it is not rare to encounter different works addressing the same task but using a different name. Here, we try to overcome this limitation and aim to unify the nomenclature for open-world tasks, hoping that this would make open-world segmentation more common and accessible in the future.

In summary, our contributions are threefold. First, we propose a clear and unified nomenclature for open-world segmentation, resolving the inconsistencies that arise when different works use different terms to refer to the same task. Second, we introduce a fully convolutional neural network that, through a novel pre-logit loss formulation, achieves compelling performance on open-world segmentation tasks while retaining strong closed-world performance. In contrast to prior segmentation-based category discovery methods that operate offline, our approach incrementally discovers novel categories at test time and maintains cross-image category consistency through online updates of semantic descriptors. Third, we present PANIC, a dataset of 800 autonomous-driving images with pixel-wise annotations covering more than 50 anomalous categories and over 4000 object instances, accompanied by public challenges for multiple open-world segmentation tasks. 

  \section{Related Work}\label{sec:related}

Recently, semantic and panoptic segmentation have been among the most prominent perception tasks in different domains, such as autonomous driving~\citep{borse2021cvpr, roynard2018ijrr, wang2023cvpr}, indoor navigation~\citep{hu2021cvpr, kundu2020eccv, schwarz2018ijrr, sodano2023icra}, and agricultural robotics~\citep{guclu2025ijrr, halstead2024ijrr, marks2023ral, roggiolani2023icra}. However, they mostly operate in closed-world settings, while the open-world domain has been relatively under-explored.
 
\textbf{Open-World Detection and Segmentation}.
The first approaches that relaxed the closed-world assumption targeted anomaly detection and open-set recognition. Their primary goal was to identify and reject previously unseen samples. Early methods adapted standard closed-world segmentation pipelines by introducing dedicated anomaly classes~\citep{blum2021ijcv, mor2018wacv, vaze2021iclr}, modifying or thresholding the softmax output~\citep{bendale2016cvpr, cardoso2017ml, hendrycks2017iclr}, or using model ensembles~\citep{ji2025ijrr, lakshminarayanan2017nips, vyas2018eccv}. Although influential, those approaches often struggled to reliably identify unknown samples because deep networks tend to produce overconfident predictions even for out-of-distribution inputs~\citep{nguyen2015cvpr, tommasi2017dacva}.

To address this limitation, folllow-up works explored alternative signals for anomaly detection, including entropy maximization~\citep{dhamija2018neurips}, energy-based scores~\citep{liu2020neurips}, and uncertainty estimation through Bayesian deep learning~\citep{gal2016icml, lakshminarayanan2017nips, sapkota2022cvpr} or gradient-based measures~\citep{liu2018cvpr, maag2024visapp}. Other methods leverage examples of unknown objects during training to better separate known and unknown regions at test time~\citep{hwang2021cvpr, wong2020corl}. Generative models have also been explored for open-world perception. While some approaches detect unknowns through reconstruction errors~\citep{zhao2023cvpr-oauc}, others, such as OpenGAN~\citep{kong2021cvpr}, generate synthetic open-set examples to improve the discrimination between known and unknown samples. More recently, attention has shifted toward open-set panoptic segmentation, where anomalous regions are further decomposed into object instances using class-agnostic grouping strategies~\citep{gasperini2023iccv, nayal2023iccv, nekrasov2023gcpr, rai2024tpami, schmidt2025iccv}. The ability to segment previously unseen objects is closely related to the literature on object proposals and class-agnostic instance segmentation. DeepMask~\citep{pinheiro2015neurips} learns category-independent object proposals by exploiting generic visual cues such as object boundaries and shape regularities, while Recurrent Pixel Embedding~\citep{kong2018cvpr} groups pixels into instances through learned embedding representations without relying on category-specific predictions. These works suggest that object-centric cues can generalize beyond the training taxonomy and provide an important inductive bias for reasoning about unseen objects. Our approach builds on a similar intuition, but addresses the more challenging open-world panoptic segmentation setting, where the model must not only identify and segment unknown instances, but also discover and maintain semantic category identities across images. To encourage such object-centric behavior, we further incorporate geometric priors through divergence- and curl-based regularization of the learned offset field.

A related but less explored problem is the discovery of multiple novel categories. In semantic segmentation, NCDSS~\citep{zhao2022cvpr} discovers novel semantic classes from unlabeled images containing the categories to be learned, assuming that the number of unknown categories is known beforehand. At the instance level, \cite{weng2021cvpr} and RNCDL~\citep{fomenko2022neurips} jointly perform unknown instance segmentation and category discovery, producing cross-image consistent labels for novel categories. However, these methods operate in an offline setting, where the entire dataset is available before category discovery takes place.

The online counterpart of this problem has been studied in image classification through online category discovery. Methods such as OCD~\citep{du2023cvpr} and PHE~\citep{zheng2024neurips} process samples sequentially and decide whether each incoming observation belongs to a previously known class or to a newly discovered category. Open-world semantic segmentation extends this idea to dense prediction by incrementally discovering multiple semantic categories while maintaining category consistency across images~\citep{sodano2024cvpr}. Open-world panoptic segmentation further extends this setting by additionally requiring instance-level segmentation. As a result, it combines semantic category discovery, unknown instance segmentation, and online category formation within a single framework.

Another related research direction is open-vocabulary segmentation, also known as zero-shot segmentation. Those methods aim to recognize and segment categories that were not seen during training by leveraging large-scale vision-language models such as CLIP~\citep{radford2021icml}. Most approaches rely on text prompts or predefined vocabularies to guide the segmentation process~\citep{ghiasi2022eccv, liang2023cvpr, qin2023cvpr, xu2023iccv, zhang2023iccv-asff}, requiring potential object categories to be specified beforehand. More recently, promptless VLM-based approaches such as Zero-Guidance Segmentation~\citep{rewatbowornwong2023iccv} and AutoSeg~\citep{ulger2025iccv} have explored automatic category labeling without user-provided prompts by combining class-agnostic segmentation with vision-language representations. While these methods remove the need for predefined vocabularies, the resulting labels remain image-dependent and do not explicitly maintain category identities across images. Consequently, semantically equivalent objects may receive different textual descriptions or be represented at different levels of granularity. In contrast, open-world panoptic segmentation focuses on discovering and maintaining consistent category identities across images while simultaneously providing semantic and instance-level segmentation, without relying on language supervision at inference time.

In this paper, we present a fully convolutional neural network for open-world panoptic segmentation. Our goal is to discover multiple unknown semantic categories while simultaneously segmenting their individual object instances. Unlike prior approaches, we do not rely on unknown objects during training, predefined vocabularies, or vision-language supervision. To the best of our knowledge, our approach is the first method specifically designed for open-world panoptic segmentation, jointly addressing online category discovery and unknown instance segmentation.

\begin{figure*}[t!]
    \centering
    \includegraphics[width=\linewidth]{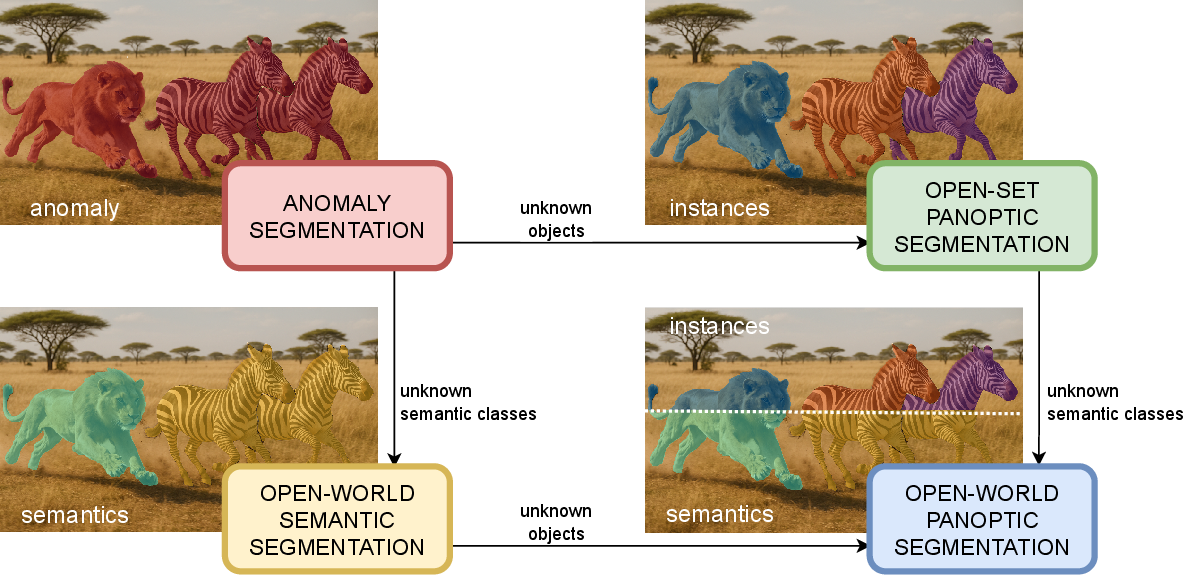}
    \caption{A visual breakdown of the four open-world segmentation tasks. Anomaly segmentation segment all anomalous areas as unknown (zebras and lion together). Open-world semantic segmentation separates classes but has no objects (zebras segmented together, lion separate). Open-set panoptic segmentation segments separate objects but has no category information. Open-world panoptic segmentation has both, classes and object information. Exemplary RGB image is generated with perplexity.ai~\citep{perplexity2025}. \vspace{-1em}}
    \label{fig:task_breakdown}
\end{figure*}

\textbf{Datasets}.
While large-scale datasets exist for closed-world segmentation in different domains~\citep{behley2021ijrr, cordts2016cvpr, lin2014eccv, sethuraman2025ijrr, vidanapathirana2025ijrr, weyler2023tpami}, datasets for open-world segmentation are comparably rare. The WildDash dataset~\citep{zendel2018eccv} provides anomalous images where full-image anomalies are present. Another dataset is MVTec~\citep{bergmann2019cvpr}, which mostly targets the industrial scenario. BrainMRI \& HeadCT~\citep{salehi2021cvpr} are two datasets targeting the medical domain for detecting lesions on different organs. ALLO~\citep{leveugle2024arxiv} is a synthetic photorealistic dataset with full-image anomalies for robot operations in lunar orbit. Among the most common datasets for open-world segmentation in the autonomous driving domain are the Fishyscapes-LostAndFound benchmark~\citep{blum2019cvprws,pinggera2016iros}, which is based on the same setup as Cityscapes and presents anomalous objects in the middle of the street, the CAOS benchmark~\citep{hendrycks2022icml} that originates from the BDD100K~\citep{yu2020cvpr} dataset to create an open-world test set, and TAO-OW~\citep{liu2022cvpr} for open-world instance tracking. These datasets are, however, characterized by a low diversity of anomalies. \cite{chan2021neurips} introduced the SegmentMeIfYouCan benchmark, which is a test set that relies on the known classes from Cityscapes and introduces anomalies of various kinds and sizes. However, it provides a limited number of images, and no semantic or instance annotation, but only a binary segmentation mask between known and unknown. Recently, \cite{nekrasov2024arxiv} extended SegmentMeIfYouCan by adding instance information, even though semantic information is still missing. 

In this work, we propose a dataset for open-world panoptic segmentation with more than 50 unknown classes and more than 4000 instances. We assume everything that appears in Cityscapes to be known, and provide pixel-wise annotations for the unknowns. We also provide public competitions for several open-world segmentation task, spanning from anomaly segmentation to open-world panoptic segmentation.

\section{Task Description}\label{sec:tasks}

Open-world segmentation remains a relatively under-explored area, having gained momentum only in recent years with the introduction of dedicated benchmarks~\citep{blum2019cvprws, chan2021neurips, nekrasov2024arxiv} and with adaptations of closed-world datasets for open-world evaluation~\citep{hendrycks2022icml, lin2014eccv}. This emerging status has led to terminology that is often ambiguous or inconsistently used across the literature. For example, the task of discovering multiple novel semantic categories has been referred to as both open-set and open-world semantic segmentation~\citep{rai2024tpami, sodano2024cvpr}. Likewise, the task of identifying a single ``unknown'' super-category and segmenting all object instances within it has appeared under names such as open-world instance segmentation~\citep{nunes2022ral}, open-set panoptic segmentation~\citep{rai2024tpami}, and open-world panoptic segmentation~\citep{chakravarthy2025ijcv, schmidt2025iccv}. Additionally, some works use the term open-set semantic segmentation~\citep{kong2021cvpr} to describe the task of assigning pixels either to one of the known semantic classes or to a generic unknown category. In our formulation, this setting is equivalent to anomaly segmentation, since all tasks considered build upon closed-world semantic segmentation and therefore always predict the known classes alongside any unknown regions. 

Thus, in this section, we propose a clear and unified terminology for all open-world segmentation tasks and use it throughout the paper. In~\figref{fig:task_breakdown}, we show a visual example of the four tasks on an exemplary image that has been manually labeled for illustration purposes. All tasks build on top of closed-world segmentation, i.e., they output prediction masks for known classes as well. Additionally, none of these tasks relies on a known vocabulary of categories to be discovered, as open-vocabulary approaches are not covered.

\subsection{Anomaly Segmentation} \label{subsec:an_seg}
SegmentMeIfYouCan~\citep{chan2021neurips} first formalized the task of \textit{anomaly segmentation}: given a model trained on a certain number of known categories, anomaly segmentation segments the areas belonging to none of them. There is no concept of unknown category or object. The input image is segmented into two areas: one belonging to everything that is known (i.e., seen at training time), and one including everything else. 

Given an image $\b{I} \in \mathbb{R}^{H \times W}$, the prediction for anomaly segmentation is a mask $\b{M}_a \in \{0, 1\}^{H \times W}$, where $0$ indicates known and $1$ indicates unknown.

\subsection{Open-World Semantic Segmentation} \label{subsec:ow_semseg}
Open-world semantic segmentation~\citep{sodano2024cvpr} segments the anomalous area into different semantic categories. As for standard semantic segmentation, these categories stay consistent (i.e., areas belonging to the same category across different images are assigned to the same semantic label). This task has no object instances. 

Given an image $\b{I} \in \mathbb{R}^{H \times W}$, the prediction for this task is a mask $\b{M}_s \in \{0, \, \dots, \, K\}^{H \times W}$, with $0$ for known and $1, \, \dots, \, K$ for the $K$ new categories.

\subsection{Open-Set Panoptic Segmentation} \label{subsec:os_panseg}
Open-set panoptic segmentation is more common in the literature~\citep{hwang2021cvpr, rai2024tpami}, and consists of segmenting individual objects within the anomaly segmentation mask. This task does not look at finding new semantic classes, but rather new objects. Generally speaking, this task performs class-agnostic instance segmentation within the anomalous area. 

Given an image $\b{I} \in \mathbb{R}^{H \times W}$, the prediction for this task is a mask $\b{M}_a \in \{0, 1\}^{H \times W}$, where $0$ indicates known and~$1$ indicates unknown, and a mask \mbox{$\b{M}_i \in \{0, \, \dots, \, N\}^{H \times W}$} where $0$ indicates both known and no object areas, and $1, \, \dots, \, N$ indicate the $N$ object instances discovered.

\begin{figure*}[t!]
    \centering
      \includegraphics[width=.9\textwidth]{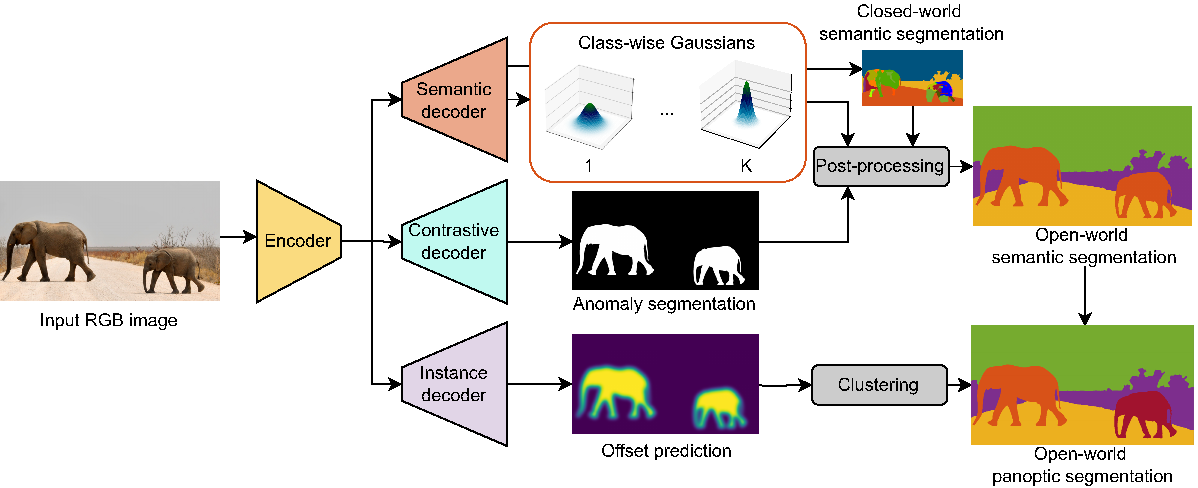}
      \caption{Our network processes an RGB image via an encoder and three decoders, for semantic segmentation, anomaly segmentation and class-agnostic instance segmentation. The semantic segmentation decoder also builds class descriptors for the known categories. Results are post-processed and yield the final open-world panoptic segmentation result. \vspace{-1em}}
      \label{fig:architecture}
    \end{figure*}

\subsection{Open-World Panoptic Segmentation} \label{subsec:ow_panseg}
Open-world panoptic segmentation builds on open-world semantic segmentation and aims to discover not only novel and consistent semantic classes, but also individual objects within these classes. It goes a step further from open-set panoptic segmentation as well, since the latter does not aim to discover new semantic classes. The task of open-world panoptic segmentation gives a holistic understanding of the scene, since every pixel belongs to a semantic class and will have an instance ID (when applicable), irrespectively of it being \textit{known} or \textit{unknown}. 

Given an image $\b{I} \in \mathbb{R}^{H \times W}$, the prediction for open-world panoptic segmentation is a mask $\b{M}_s \in \{0, \, \dots, \, K\}^{H \times W}$, where $0$ indicates known and $1, \, \dots, \, K$ indicate the $K$ categories discovered, and a mask $\b{M}_i \in \{0, \, \dots, \, N\}^{H \times W}$, where~$0$ indicates both known and no object areas, and~$1, \, \dots, \, N$ indicate the $N$ object instances discovered.

Open-world panoptic segmentation provides a richer scene representation than its related variants because it jointly discovers previously unseen semantic categories and their individual object instances, without requiring prior prompts or textual descriptions. Unlike open-world semantic segmentation, which identifies unknown regions but does not separate individual objects, open-world panoptic segmentation enables reasoning at the instance level. This allows an autonomous system to estimate how many unknown objects are present and to distinguish them individually. Compared to open-set panoptic segmentation, which groups all unknown pixels into a single category, it further enables the discovery of meaningful structure among previously unseen objects.

While binary anomaly segmentation may be sufficient for certain safety-critical behaviors such as triggering a stop or slowdown, many downstream robotics tasks are inherently object-centric. Multi-object tracking, behavior prediction, and planning often require reasoning about individual agents rather than undifferentiated anomaly regions. In this context, distinguishing and grouping unknown objects may provide additional information that can support future decision-making processes. More generally, we view open-world panoptic segmentation as a step toward holistic scene understanding in environments where not all object categories are known during training, extending the panoptic segmentation paradigm to open-world settings.

In this article, we propose a novel method and provide a benchmark for open-world panoptic segmentation.

\section{Our Methodology}\label{sec:main}

We address the problem of open-world panoptic segmentation. Our approach, illustrated in~\figref{fig:architecture}, uses a fully convolutional neural network with a single encoder and three decoders. Since our contributions focus primarily on the design of loss functions and on the strategy used to discover novel categories and objects, the method is not tied to this particular architectural choice. In fact, it can be integrated into any other backbone network without modification.

\subsection{Encoder Architecture} \label{subsec:network_arch}
We use a lightweight fully-convolutional neural network, to facilitate future deployment in the real world. Our CNN is composed of one encoder and three decoders. We use a ResNet34~\citep{he2016cvpr} encoder, where we replaced the standard ResNet block with the NonBottleneck-1D block~\citep{romera2018tits}. This choice is motivated solely from the fact that we aim to make our architecture lighter, and the NonBottleneck-1D block replaces all $3 \times 3$ convolution by a sequence of $3 \times 1$ and $1 \times 3$ convolutions with a ReLU in between. Additionally, we incorporate contextual information by means of a pyramid pooling module~\citep{zhao2017cvpr-pspn} after the encoder. 

\subsection{Decoder Architectures} \label{subsec:approach}
Our three decoders have the same structure and are composed of three SwiftNet modules~\citep{orsic2019cvpr} with NonBottleneck-1D blocks, and two final upsampling modules based on nearest-neighbor and depth-wise convolutions~\citep{chollet2017cvpr}, that have the major advantage of substantially reducing the computations needed. Following the standard U-Net~\citep{ronneberger2015miccai} design, we also employ encoder-decoder skip connections after each downsampling stage of the encoder to propagate fine-grained features to the three decoders. 

The first decoder, called ``semantic decoder'', targets semantic segmentation and additionally builds a class descriptor for each known category. At inference time, we compare the final features to the known descriptors to discriminate between known and unknown classes, building descriptors for unknown classes as well. The second decoder, called ``contrastive decoder'', targets anomaly segmentation by mapping the features of pixels belonging to known classes on a unit hypersphere, and the ones belonging to unknown classes to $0$. The third decoder, called ``instance decoder'' targets class-agnostic instance segmentation by means of vector fields-inspired loss functions~\citep{weyler2024ral}. The outputs of the decoders go through different post-processing phases to obtain the final open-world panoptic segmentation result.

Consider an image \mbox{$\b{I} \in \mathbb{R}^{H \times W}$}. We refer to the set of pixels in the image as \mbox{$\Omega = \{(1, \, 1), \, \dots, \, (H, \, W)\}$}. Additionally, we call \mbox{$\b{Y} \in \{1, \, \dots, \, K\}^{H \times W}$} the ground truth semantic mask of $\b{I}$, where $K$ is the number of known categories used at training time. In contrast, we refer to the predicted (closed-world) semantic mask as \mbox{$\hat{\b{Y}} \in \{1, \, \dots, \, K\}^{H \times W}$}. Similarly, we call $\b{Z}$ the ground truth and $\hat{\b{Z}}$ the predicted instance mask, respectively. Finally, we denote with \mbox{$\Omega_k = \{p \in \Omega \, \mid \, Y_p = k\}$} the set of pixels in the image whose ground truth label is $k$, and with \mbox{$\hat{\Omega}_k = \{p \in \Omega \, \mid \, \hat{Y}_p = Y_p = k\}$} the set of pixels whose predicted label $\hat{\b{Y}}_p$ and ground truth label~$\b{Y}_p$ are both equal to~$k$, i.e., the set of true positives for class~$k$. We do not use bold notation for the pixel tuple~$p$ for the sake of clarity.

\textbf{Semantic Decoder}. The semantic decoder targets semantic segmentation and it leverages the standard weighted cross-entropy loss:
\begin{equation}
  \mathcal{L}_{\mathrm{sem}} = - \dfrac{1}{|\Omega|} \sum_{p \in \Omega} \omega_k \, \b{t}_p^\top \, \log \big( \sigma(\b{f}_p) \big),
  \label{eq:ce_loss}
\end{equation}
where $\omega_k$ is a class-wise weight computed via the inverse frequency of each class in the dataset, \mbox{$\b{t} \in \mathbb{R}^{H \times W \times K}$} is the one-hot encoded ground truth annotation of the image, $\b{t}_p \in \mathbb{R}^K$ is the one-hot encoded ground truth annotation at pixel location $p$, $\sigma (\cdot)$ denotes the softmax operation, and $\b{f}_p$ denotes the pre-softmax feature predicted at pixel $p$.

Unlike approaches that operate on pre-softmax features~\citep{sodano2024cvpr}, whose dimensionality is tied to the number of known classes and can therefore limit performance in few known classes scenarios, we apply our loss directly to the higher-dimensional, class-agnostic pre-logit features $\logit_p$, where $p$ indicates the pixel. This allows us to learn more expressive and discriminative class prototypes.

To build the class descriptors, during training we accumulate the pre-logits of all true positives for each class, where a true positive is a pixel that is correctly segmented. By doing this, we can build a running average class descriptor $\b{\mu}_k \in \mathbb{R}^D$ for each class \mbox{$k \in \{1, \, \dots, \, K\}$} that appears at training time:
\begin{equation}
  \b{\mu}_k = \dfrac{1}{|\hat{\Omega}_k|} \sum_{p \in  \hat{\Omega}_k} \logit_p.
  \label{eq:mean}
\end{equation}

Together with the mean, we also compute the per-class variance $\b{\sigma}^2_k \in \mathbb{R}^D$ via the sum of squares:
\begin{equation}
  \b{\sigma}^2_k = \dfrac{1}{|\hat{\Omega}_k|} \sum_{p \in  \hat{\Omega}_k} (\logit_p - \b{\mu}_k) \odot (\logit_p - \b{\mu}_k),
  \label{eq:variance}
\end{equation}
where $\odot$ indicates the element-wise product (Hadamard product). 

At the beginning of epoch $e$, each category has a mean $\b{\mu}^{e-1}_k$ and a variance $\b{\sigma}^{e-1}_k$, computed at the previous epoch $e-1$. The goal of having a unique class descriptor for each known class is that the prediction at each pixel belonging to that class must be equal to the descriptor. Thus, we use the feature loss function $\mathcal{L}_{\mathrm{feat}}$ at the pre-logit level given by
\begin{equation}
  \mathcal{L}_{\mathrm{feat}} = \dfrac{1}{|\Omega|} \sum_{k = 1}^K \sum_{p \in \Omega_k} \dfrac{\Big|\Big| \logit_p - \b{\mu}^{e-1}_k \Big|\Big|^2}{\big( \b{\sigma}^{e-1}_k \big)^\top \big( \b{\sigma}^{e-1}_k \big)}.
  \label{eq:feat_loss}
\end{equation}

The idea is that true positives help build class descriptors, as specified in~\eqref{eq:mean}, which are then used as supervision in the next epoch. This loss formulation extends our previous work~\citep{sodano2024cvpr} by operating on the pre-logit representation rather than the pre-softmax feature. This removes the dependence of the feature dimensionality on the number of known training classes, making the representation more suitable for category discovery when only few known categories are available.

The semantic decoder is optimized with a weighted sum of the loss functions introduced above:
\begin{equation}
  \mathcal{L}_{\mathrm{sdec}} = w_1 \, \mathcal{L}_{\mathrm{sem}} + w_2 \, \mathcal{L}_{\mathrm{feat}}.
\end{equation}

\textbf{Contrastive Decoder}. The contrastive decoder targets anomaly segmentation, and aims to predict features whose norm is close to $1$ for pixels belonging to known classes, and to $0$ for pixels belonging to unknown classes. 

This decoder also operates at the pre-logit level and combines the contrastive loss~\citep{chen2020icml} and the objectosphere loss~\citep{dhamija2018neurips}. To do so, we first compute the mean pre-logit $\bar{\b{\ell}}_k \in \mathbb{R}^D$ for each class $k$ in the current image from the pre-logit predicted by this decoder. We compute the contrastive loss $\mathcal{L}_{\mathrm{cont}}$ such that $\bar{\b{\ell}}_k$ approximates the normalized mean representation $\bar{\b{\mu}}_k^{e-1}$ of the corresponding class in the previous epoch $\b{\mu}_k^{e-1}$ and gets dissimilar from the other classes mean representation:
\begin{equation}
  \mathcal{L}_{\mathrm{cont}} = - \sum_{k=1}^{K} \log \frac{\exp{({\overline{\b{\ell}}_k}^\top \bar{\b{\mu}}_{k}^{e-1} / \tau})}{\sum_{i=1}^K{\exp{({{\overline{\b{\ell}}_k}^\top \bar{\b{\mu}}_i^{e-1} / \tau})}}},
  \label{eq:closs}
\end{equation}
where $\tau$ is a temperature parameter. The goal is to have~$\bar{\b{\ell}}_k$ approximate the corresponding vector in the feature bank, i.e., the normalized class descriptors $\bar{\b{\mu}}_k^{e-1}$. To achieve this, we changed from the pre-softmax features to the pre-logit in this loss function, in order to have dimensionally-consistent vectors. The outcome of the contrastive loss is to scatter the class-specific pre-logits on the unit hypersphere.
Additionally, we use the objectosphere loss function
\begin{equation}
  \mathcal{L}_{\mathrm{obj}} = \left\{ \begin{array}{cl}
    \max \big( 1 - \norm{\b{\ell}_{p}}^2, \, 0 \big) &, \mathrm{if}  \, p \in \Omega_{k} \\ 
    \norm{\b{\ell}_{p}}^2  &, \mathrm{otherwise} \end{array}\right.,
  \label{eq:obj_loss}
\end{equation}
where $\Omega_{k}$ is the set of pixels in the image belonging to known classes, while the others belong to unlabeled (void) areas of the image and are treated as unknown. In practice, these void regions correspond to pixels annotated as ``void'' in Cityscapes and are used as known-unknown examples during training.
The loss pushes known pre-logits to have norm $\geq 1$ and unknowns to have norm $0$. Although other values could be used, norm $1$ aligns with the contrastive loss in~\eqref{eq:closs}, which maps features to the unit sphere. Using a different norm would conflict with this goal, especially values larger than $1$.

We optimize the contrastive decoder with a weighted sum of the loss functions introduced above:
\begin{equation}
  \mathcal{L}_{\mathrm{cdec}} = w_3 \, \mathcal{L}_{\mathrm{cont}} + w_4 \, \mathcal{L}_{\mathrm{obj}}.
\end{equation}

The contrastive decoder is used for anomaly segmentation and could, in principle, be replaced by alternative architectures serving the same purpose. We adopt this design because it is lightweight, architecturally consistent with the semantic decoder, and operates in the same pre-logit representation space used for category discovery.

\textbf{Instance Decoder}. The instance decoder addresses class-agnostic instance segmentation, where we predict a 2D offset for each pixel, so that pixels belonging to the same object instance point to the same area in the image and can be clustered together.

Given an instance $\mathcal{C}_j \subset \Omega$, defined as the set of pixels of the images belonging to the $j$-th object instance, with a centroid $\b{c}_j$, we want to obtain offset vectors for all pixels belonging to $\mathcal{C}_j$ that point toward $\b{c}_j$, while the remaining offset vectors should not. We achieve this by means of the Lov\'asz Hinge loss~\citep{neven2019cvpr}
\begin{equation}
  \mathcal{L}_{\mathrm{off}} = \dfrac{1}{|\mathcal{C}|} \sum_{j=1}^{|\mathcal{C}|} \mathrm{Lov\acute{a}sz} (\b{F}_{\mathcal{C}_j}, \, \b{G}_{\mathcal{C}_j}).
  \label{eq:offset_loss}
\end{equation}

In this equation, $\mathcal{C}$ is the set of instances in the image, $\b{F}_{\mathcal{C}_j} \in \mathbb{R}^{H \times W}$ is a soft-mask obtained from the offset predictions for optimization purposes, while $\b{G}_{\mathcal{C}_j} \in \{0, \, 1\}^{H \times W}$ denotes the binary ground truth mask of the $j$-th instance. The soft mask $\b{F}_{\mathcal{C}_j}$ is not directly predicted by the network; instead, it is an intermediate quantity derived from the predicted offsets and used exclusively within the Lovász-Hinge loss. Specifically, each pixel receives a score based on how close the location pointed to by its predicted offset is to the centroid of instance $\mathcal{C}_j$. This is formalized as
\begin{equation}
  f_{\mathcal{C}_j} = \exp \bigg( -\dfrac{||\b{e}_p - \b{c}_j||^2}{2 \eta^2} \bigg),
\end{equation}
where $\b{e}_p = p $ indicates the pixel location pointed by the prediction at pixel $p$, and $\eta$ is a hyperparameter that defines an isotropic clustering region around the centroid. 
This loss function encourages the network to predict, for all pixels associated with a specific instance, an offset vector that points toward their corresponding centroid. Simultaneously, it penalizes offset vectors that point towards a different centroid than their own. This forms distinct clusters in 2D space, even if the centroid lies outside the instance.

Following the idea formulated by \cite{weyler2024ral}, we adopt divergence and curl loss functions to refine offset predictions. These two loss functions are inspired by the concept of vector field in physics, and have so far only been applied for closed-world segmentation in the case of partial annotations. Here, we show their effectiveness for open-world segmentation, and provide a quantitative analysis of each component in the ablation study.

Divergence captures how pixels ``flow'' toward or away from a point, aligning well with the idea of pixels pointing to a centroid. Curl measures rotation in the field, helping to regularize the vector pattern.
For formalizing the divergence and curl loss functions, we need a simplified vector field defined by a continuous multivariable function \mbox{$\b{O}(h, \, w) \, : \, \mathbb{R}^2 \rightarrow \mathbb{R}^2$} given by:
\begin{equation}
  \mathbf{O}\left(h,w\right)
  =
  \begin{bmatrix}
    o_{h}\left(h,w\right) \\
    o_{w}\left(h,w\right)
  \end{bmatrix}
  =
  \begin{bmatrix}
    -h + c_h \\
    -w + c_w
  \end{bmatrix},
  \label{eq:simple_vf}
\end{equation}
that behaves as a perfect sink, i.e., all offset vectors point toward $\b{c} = \big[c_h, \, c_w\big]^\top$. In 2D, divergence is defined as
\begin{equation}
  \mathrm{div} \, \mathbf{O}:=
  \frac{\partial o_{h}\left(h,w\right)}{\partial h} 
  +
  \frac{\partial o_{w}\left(h,w\right)}{\partial w}
  .
  \label{eq:def_divergence}
\end{equation}
Thus~${\frac{\partial o_{h}\left(h,w\right)}{\partial h} =
\frac{\partial o_{w}\left(h,w\right)}{\partial w} = -1}$ and~${\mathrm{div}\, \mathbf{O}}=-2$. We enforce that divergence is equal to $-2$ and its two components must be equal with two regression loss functions:
\begin{equation}
  \begin{split}
  \mathcal{L}_\mathrm{div} & =
  \frac{1}{|\Omega|}
  \sum_{p \in \Omega}^{}
  \rho
  \left(
    \left(\mathrm{div} \, \mathbf{O}\right)_{p} - (-2)
    \right) \\
    \mathcal{L}^{\mathrm{aux}}_\mathrm{div} & =
    \hspace{-0.05cm}
    \frac{1}{|\Omega|}
    \hspace{-0.1cm}
    \sum_{\left(h,w\right) \in \Omega}^{}
    \hspace{-0.25cm}
    \rho
    \left(
      \hspace{-0.05cm}
      \frac{\partial o_{h}\left(h,w\right)}{\partial h}
      -
      \frac{\partial o_{w}\left(h,w\right)}{\partial w}
      \hspace{-0.05cm}
      \right),
  \end{split}   
\end{equation}
where $\rho(\cdot)$ denotes the Geman-McClure loss~\citep{barron2019cvpr}, and $(h, \, w)$ denote the image coordinates of pixel $p$.

The curl of a vector field $\mathbf{O}$ is defined as 
\begin{equation}
  \mathrm{curl} \, \mathbf{O} :=
  \frac{\partial o_{h}\left(h,w\right)}{\partial w} 
  -
  \frac{\partial o_{w}\left(h,w\right)}{\partial h}.
  \label{eq:def_curl}
\end{equation}
When applying the curl to the vector field in~\eqref{eq:simple_vf}, we obtain ${\frac{\partial
o_{h}\left(h,w\right)}{\partial w} = \frac{\partial
o_{w}\left(h,w\right)}{\partial h} = 0}$, which imply \mbox{${\mathrm{curl} \, \mathbf{O}}=0$}. Concretely speaking, this means we want no rotational behavior in our vector field. We formalize two loss functions to have the curl equal to zero and the two partial derivatives to be equal to each other:
\begin{equation}
  \begin{split}
  \mathcal{L}_\mathrm{curl} & =
  \frac{1}{|\Omega|}
  \sum_{p \in \Omega}^{}
  \rho
  \left(
    \left(\mathrm{curl} \, \mathbf{O}\right)_{p}
    \right) \\
  \mathcal{L}^{\mathrm{aux}}_\mathrm{curl} & =
    \hspace{-0.05cm}
    \frac{1}{|\Omega|}
    \hspace{-0.1cm}
    \sum_{\left(h,w\right) \in \Omega}^{}
    \hspace{-0.25cm}
    \rho
    \left(
      \hspace{-0.05cm}
      \frac{\partial o_{h}\left(h,w\right)}{\partial w}
      -
      \frac{\partial o_{w}\left(h,w\right)}{\partial h}
      \hspace{-0.05cm}
      \right).
  \end{split}
\end{equation}

The instance decoder is optimized with the weighted sum of all losses introduced so far
\begin{equation}
  \begin{split}
  \mathcal{L}_{\mathrm{ins}} = & w_5 \, \mathcal{L}_{\mathrm{off}} + w_6 \,\mathcal{L}_{\mathrm{div}} + w_7 \, \mathcal{L}_{\mathrm{div}}^{\mathrm{aux}} + \\ & w_8 \,\mathcal{L}_{\mathrm{curl}} + w_9 \, \mathcal{L}_{\mathrm{curl}}^{\mathrm{aux}}.
  \end{split}
\end{equation}

\subsection{Post-Processing}
Given the outputs of the three decoders, we now describe the generation of the corresponding masks.

\textbf{Post-Processing for Anomaly Segmentation}. We obtain anomaly segmentation results by fusing the outputs of the semantic and contrastive decoders. The semantic decoder provides class descriptors (mean $\b{\mu}_k$ and variance $\b{\sigma}_k^2$) in the pre-logit space for each known class, allowing us to construct a multi-variate normal distribution $\mathcal{N}_k(\b{\mu}_k, \, \b{\Sigma}_k)$, where $\b{\Sigma}_k = \mathrm{diag}(\b{\sigma}^2)$ is the covariance matrix, which we obtain from the variance assuming that all classes are independent. In this way, for each predicted pre-logit $\b{\ell}_p$ at pixel $p$, we can compute its fitting score to each known category by means of the squared exponential kernel
\begin{equation}
  s_k(\b{\ell}_p) = \exp \bigg(- \dfrac{1}{2} (\b{\ell}_p - \b{\mu}_k)^\top \b{\Sigma}_k^{-1} (\b{\ell}_p - \b{\mu}_k) \bigg).
  \label{eq:sek}
\end{equation}

Then, we take the highest score 
\begin{equation}
  s(p) = \max_k s_k(\b{\ell}_p),
  \label{eq:max_score}
\end{equation}
which corresponds to the closest known category to the predicted pixel. 
At test time, this process replaces the standard procedure for closed-world semantic segmentation, i.e., $\argmax(\cdot)$ on pre-softmax outputs.

\begin{table*}[t]
  \centering
  \begin{tabular}{cccccc}
    \toprule
    \multirow{3}{*}{\textbf{Dataset}} & \multicolumn{2}{c}{\textbf{Images}} & \multirow{3}{*}{\textbf{Semantic Classes}} & \multirow{3}{*}{\textbf{Instances}} & \multirow{3}{*}{\textbf{Hidden Test Set}} \\
    \cmidrule(lr){2-3}
    & Val & Test & & & \\
    \midrule
    Fishyscapes Lost-and-Found~\citep{blum2019cvprws} & 373 & 1203 & N.A. & 1864 & \cmark \\
    CAOS BDDAnomaly~\citep{hendrycks2022icml} & 0 & 810 & 3 & 1231 & \xmark \\
    RoadObstacle21~\citep{chan2021neurips} & 0 & 327 & N.A. & 388 & \cmark \\
    SegmentMeIfYouCan~\citep{chan2021neurips} & 10 & 100 & N.A. & 262 & \cmark \\
    \midrule
    PANIC (ours) & 131 & 679 & 58 & 4029 & \cmark \\
    \bottomrule
  \end{tabular}
  \caption{Comparison of open-world segmentation datasets. In the semantic classes, ``N.A.'' means that there is no label.}
  \label{tab:dataset_comparison}
  \vspace{-1.5em}
\end{table*} 

To determine whether a pixel belongs to a known or unknown region, we apply a simple post-processing rule to the outputs of the first two decoders. For the semantic decoder, we compare each pre-logit feature to the distribution of its corresponding known class: if the feature lies within~$1\sigma$ from the class mean, the pixel is considered known; otherwise, it is marked as unknown, yielding a binary score \mbox{$s_{\mathrm{unk}, \,p}^{\mathrm{sem}} \in \{0,1\}$}. The contrastive decoder provides a complementary viewpoint by enforcing known-class features to have norm close to $1$ and unknown-class features to have norm near $0$. Under the assumption that unknown features follow a standard normal distribution $\mathcal{N}(0,1)$, known features naturally emerge as outliers. Applying the same $1\sigma$ threshold to the feature norms produces a second boolean score \mbox{$s_{\mathrm{unk},\,p}^{\mathrm{con}} \in {0,1}$}. Together, these two signals provide a robust estimation of unknown regions.

Finally, we consider a pixel to be unknown only if both $s_{\mathrm{unk}, \, p}^{\mathrm{sem}} = 1$ and $s_{\mathrm{unk}, \, p}^{\mathrm{con}} = 1$. 

\textbf{Post-Processing for Open-World Semantic Segmentation}. At test time, we have $K$ pre-logit descriptors $\b{\ell}_k$ for the known classes. For each pixel, we compare its predicted pre-logit descriptor $\b{\ell}_p \in \mathbb{R}^D$ against all existing descriptors using~\eqref{eq:max_score}. If the closest descriptor belongs to a known class and satisfies the similarity criterion, the pixel is assigned to that class. If no existing descriptor is sufficiently similar, the pixel is considered to belong to a newly discovered category and its descriptor is added to the database as a new unknown class. As described in the open-world semantic segmentation task, we aim to maintain category consistency across images. Therefore, when a pixel matches the descriptor of a previously discovered unknown category, it is assigned to that category rather than creating a new one. In this way, our database contains $\bar{K}=K+K_{\mathrm{unk}}$ descriptors, where $K_{\mathrm{unk}}$ is the number of unknown categories discovered so far, effectively treating discovered categories as ``newly known'' classes.

While descriptors of known classes remain fixed after training, descriptors of discovered unknown categories are updated online using a running mean and variance. This follows the same principle used during training for known classes, where class prototypes are updated from correctly classified samples. The difference is that, at test time, ground-truth labels are unavailable, and therefore the update is performed using all pixels assigned to the corresponding discovered category. This online update allows the category descriptors to progressively adapt as additional instances of the same unknown class are observed.

\textbf{Post-Processing for Open-World Panoptic Segmentation}. We obtain instance predictions by clustering the raw offset predictions with HDBSCAN~\citep{mcinnes2017joss}. Unlike the soft masks used during training, no thresholding or intermediate score computation is performed at inference time. We execute the clustering only in the ``thing'' areas, which means all the known ``thing'' classes, and all the unknown categories. To ensure consistency, we filter the clustering results using semantic predictions, preventing pixels from different semantic classes from being grouped into the same instance. In essence, we prioritize semantic consistency over instance grouping.

\begin{figure}[t!]
    \centering
    \includegraphics[width=.95\linewidth]{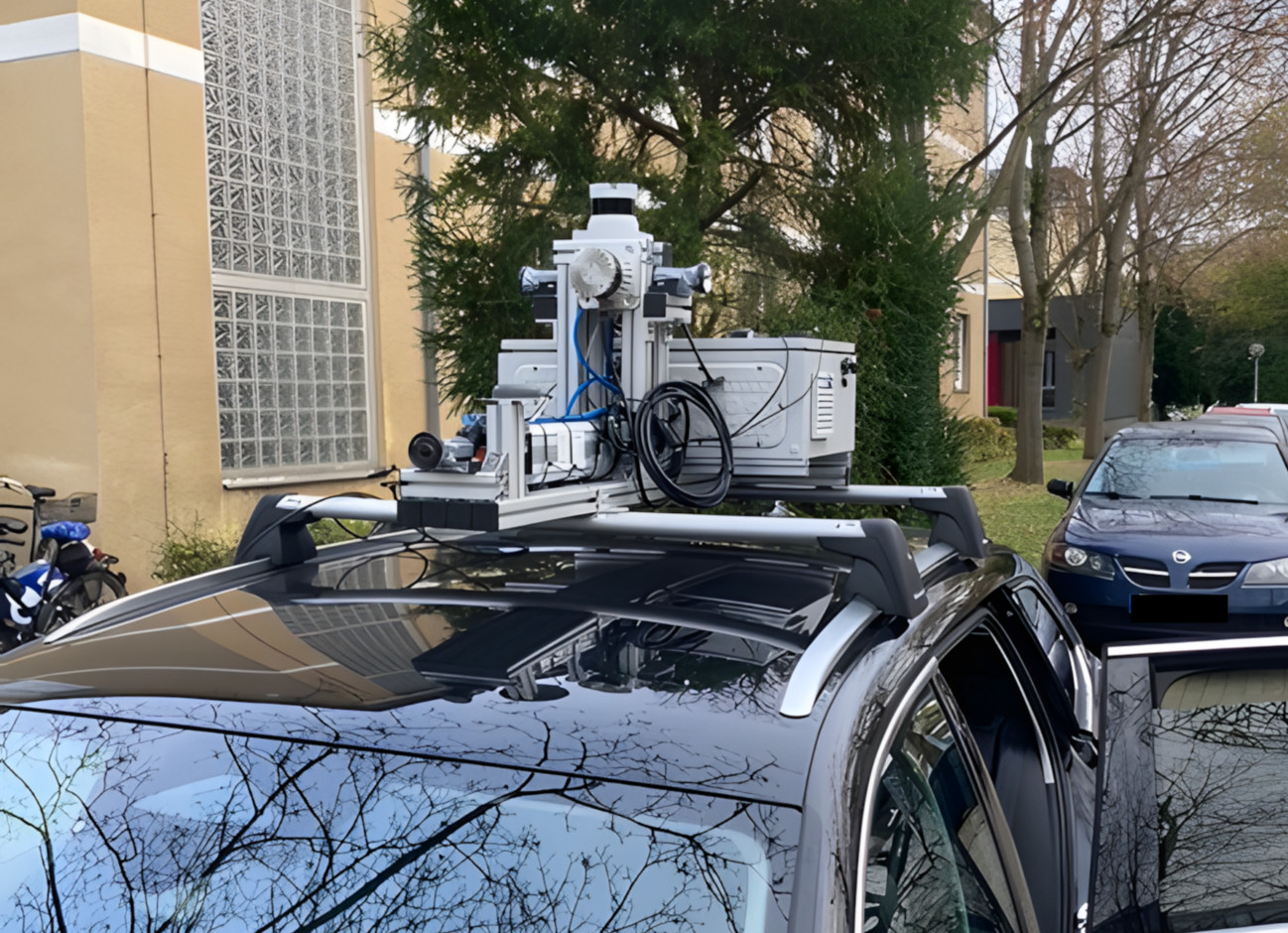}
    \caption{Sensor setup we used for recording data for the PANIC dataset. The setup includes four cameras, one GNSS/IMU device, and two 3D LiDARs. For further details, please refer to~\cite{vizzo2023itsc}. \vspace{-1em}}
    \label{fig:ipb_car}
\end{figure}

\section{The PANIC Dataset}\label{sec:panic}

In this article, we also introduce PANIC (Panoptic Anomalies In Context), an open-world panoptic segmentation test set in the autonomous driving context. Datasets for anomaly and open-world segmentation are rare compared to closed-world segmentation ones. PANIC includes 800 images, and contains 58 unknown categories. Similarly to SegmentMeIfYouCan~\citep{chan2021neurips}, we consider Cityscapes~\citep{cordts2016cvpr} as the standard training set with known classes. The 58 categories are divided into validation and test classes. 
The validation set includes classes that do not have a label to optimize for during training, but still appear at training time in the void/unlabeled area (such as garbage bins, parkimeters, monuments, etc.). In the literature, they are referred to as \textit{known unknown}~\citep{bendale2016cvpr}.  
The test set contains the \textit{unknown unknowns}, more challenging objects such as electric scooters, recumbent bicycles, forklifts, etc., that are completely absent from the training set. 

We release image and ground truths for the validation set, and only images for the hidden test set which is used for the competitions mentioned below. 

A comparison with existing datasets is shown in~\tabref{tab:dataset_comparison}. LostAndFound~\citep{blum2019cvprws} contains 9 different types of anomalies, misclassifying common Cityscapes classes like children and bicycles as anomalies. The CAOS BDD-Anomaly benchmark~\citep{hendrycks2022icml} suffers from a similar low-diversity issue, featuring just 3 unknown classes sampled from BDD100K~\citep{yu2020cvpr}. Both, Fishyscapes and CAOS, try to mitigate this low diversity by introducing synthetic data in the dataset, but this lacks real-world realism. RoadObstacle21 and SegmentMeIfYouCan~\citep{chan2021neurips} offer more diverse anomalies, but only focus on anomaly segmentation. Recently, \cite{nekrasov2024arxiv} provided instance-level annotation for these datasets, but without semantic labels, falling under open-set panoptic segmentation, where instances are detected without category differentiation.

In PANIC, anomalies can appear anywhere in the image. They can have any size, and are not limited to active traffic participants. We provide pixel-wise annotations of semantic classes and object instances.

\subsection{Data Collection and Labeling} \label{subsec:collection_labeling}
We recorded all images of PANIC in Bonn, Germany, with a car equipped with commonly used automotive sensors, such as cameras and LiDARs~\citep{vizzo2023itsc}, shown in~\figref{fig:ipb_car}. The acquisition platform additionally includes synchronized and calibrated LiDARs, IMUs, GPS receivers, and other sensors, as detailed by~\cite{vizzo2023itsc}. However, PANIC focuses exclusively on image-based open-world panoptic segmentation and therefore only utilizes RGB images acquired by the front camera, a Basler Ace acA2040-35gc. While the additional sensor modalities are not used in the present benchmark, they may provide valuable geometric information for future research on multimodal open-world perception and object discovery.

After recording, we sampled a subset of the recorded images in order to avoid duplicates and filter out images that did not contain any anomalous object, and labeled them using the online tool \href{https://segments.ai/}{segments.ai}. The 19 Cityscapes evaluation classes~\citep{cordts2016cvpr} on which most segmentation models for autonomous driving are trained, serve as basis to determine anomalies. Everything that belongs to any of those 19 classes is labeled as ``not anomaly''. For all the rest, we provide a pixel-wise semantic and instance annotation. 
To comply with privacy restrictions, we anonymize all faces, license plates, and windows of private buildings when necessary. The provided images have those areas in black. This does not bias any learning algorithm, since the dataset we propose is supposed to be an evaluation-only dataset, and the training happens on Cityscapes, which provides the known classes. Additionally, we label an ``ignore'' class that will not be included in the evaluation, as well as the anonymized areas.

We provide more details on PANIC, including the categories we labeled, in Sec.~A of the supplementary material.

\subsection{Benchmarks and Metrics} \label{subsec:bm_metrics}
We propose four benchmarks accompanied by public Codabench competitions~\citep{xu2022patterns}, one for each of the segmentation tasks described in~\secref{sec:tasks}. Each has its own evaluation pipeline and metrics.

\subsubsection{Anomaly Segmentation}
Following SegmentMeIfYouCan~\citep{chan2021neurips}, we use two groups of metrics for anomaly segmentation: pixel-level and component-level.

Pixel-level metrics include the area under the precision-recall curve (AUPR), which evaluates pixel-wise anomaly score separability and favors the minority class, and the false positive rate at $95 \%$ true positive rate (FPR95), that indicates how many false positive predictions are needed to reach~$95\%$ true positive rate. 
Pixel-level metrics, however, fail to properly evaluate small anomalies. Component-level metrics better capture all anomaly regions irrespectively of their size. SegmentMeIfYouCan~\citep{chan2021neurips} introduces three metrics for component-level evaluation: the segment-wise intersection over union (sIoU) for evaluating true positive and false negatives, the positive predicted value (PPV, or component-wise precision) for false positives, and the component-wise F1-score which summarizes true positive, false positive, and false negative. For further details, insights, and considerations, we refer to the original paper~\citep{chan2021neurips}.

We decided to not use the area under the ROC curve (AUROC), because recently several papers showed its limitations~\citep{halligan2015er, humblot2023cvpr, wang2022nips}, as two models with the same performance may differ substantially in terms of how clearly they separate in-distribution and out-of-distribution samples. In general, these works argue that AUROC is not a fair metric for comparing different approaches.

\subsubsection{Open-World Semantic Segmentation}
\label{subsubsec:owss}
Open-world semantic segmentation is still under-explored, thus a proper evaluation pipeline does not exist. Since the standard evaluation of semantic segmentation relies on intersection over union, or in short IoU~\citep{everingham2010ijcv}, we extend it to the open-world setting.

Closed-world IoU is typically computed using a confusion matrix $\b{C} \in \mathbb{R}^{K \times K}$, where $K$ is the number of known classes. Each row $i$ counts predictions for class~$i$, and a perfect IoU is achieved when only diagonal entries $(i, \, i)$ are non-zero, and matched the number of pixels belonging to ground truth class $i$. In the following, we indicate as $\mathrm{row}_i \in \mathbb{R}^K$ the $i$-th row of a confusion matrix. Similarly, $\mathrm{column}_k \in \mathbb{R}^{\tilde{K}}$ is the $k$-th column of the confusion matrix. 
The closed-world IoU concept can easily be extended to the open-world case, with two major modifications:
\begin{enumerate}
  \item The open-world confusion matrix is be square, as the evaluated approach might discover a different number of categories $\tilde{K}$ than the evaluated ones. Thus, the confusion matrix has dimension $\tilde{K} \times K$, where the number of rows $\tilde{K}$ refers to the discovered classes, and the number of columns $K$ is fixed and corresponds to the number of labeled unknown classes.
  \item The predicted class corresponding to row $i$ of the confusion matrix does not have to be matched to ground truth class $i$, but rather to $\argmax (\mathrm{row}_i)$. Simply put, we allow predicted class $i$ to match to ground truth class $j$ because, as we do not explicitly optimize for categories, we have no control over the order in which we discover new classes. 
\end{enumerate}

Notice that, in this description of the open-world IoU, we did not include known categories for simplicity of notation. Known classes would contribute  to the confusion matrix with a square block, as described above.

Beside the open-world IoU, we are also interested in understanding how reliable our predictions are. To this end, we report two clustering metrics that are particularly fitting for our case, namely homogeneity and completeness~\citep{rosenberg2007emnlp}. Homogeneity $\mathrm{Hom}_i$ measures how well each predicted cluster (i.e., discovered category) contains only samples from a single ground-truth class. It is computed as 
\begin{equation}
  \mathrm{Hom}_i = \dfrac{\max (\mathrm{row}_i)}{\sum \mathrm{row}_i}, \quad \forall \, i \, \in \{1, \, \dots, \, \tilde{K}\}.
  \label{eq:hom}
\end{equation}

Conversely, completeness $\mathrm{Com}_k$ measures the ratio of pixels of a given class that is assigned to the same cluster. In our case, it measures how consistently a certain ground truth category is predicted by the same class. It is computed as
\begin{equation}
  \mathrm{Com}_k = \dfrac{\max (\mathrm{column}_k)}{\sum \mathrm{column}_k}, \quad \forall \, k \, \in \{1, \, \dots, \, K\}.
  \label{eq:com}
\end{equation}

\subsubsection{Open-Set Panoptic Segmentation}

Open-set panoptic segmentation is evaluated by means of the panoptic quality~\citep{kirillov2019cvpr-ps}.
This evaluation procedure has been adopted by previous approaches~\citep{rai2024tpami}, and we follow it here. In this case, there are no semantic classes but only two categories, known and unknown. The unknown area contains object instances, while the known area is treated as ``stuff'', so that the quality of the closed-world instance segmentation does not affect the final panoptic quality. We also report segmentation and recognition quality. 

\subsubsection{Open-World Panoptic Segmentation}

Following standard panoptic segmentation, we split the unknown classes among \textit{things} and \textit{stuff}, and then we measure open-world mIoU, panoptic quality, recognition quality, segmentation quality, homogeneity, see~\eqref{eq:hom}, and completeness, see~\eqref{eq:com}. 

\section{Experimental Evaluation}\label{sec:exp}
The main focus of this work is an approach, Con2MAV, for open-world panoptic segmentation. We present extensive experiments on multiple datasets to show the capabilities of our method, including datasets for autonomous driving and submarine robotics applications. The results of our experiments show that our model achieves compelling results for all open-world tasks described in~\secref{sec:tasks}, while performing competitively on the known classes. Additionally, we show how our new dataset, PANIC, enables all open-world segmentation tasks and provides a comprehensive benchmark with a large number of unknown categories and object instances spanning a wide range of real-world scenarios.

\subsection{Experimental Setup} \label{subsec:exp_setup}
We use multiple datasets for validating our method, using the metrics discussed in the previous section. For anomaly segmentation, we use SegmentMeIfYouCan~\citep{chan2021neurips} and submit our prediction to the public challenge, BDDAnomaly, and our proposed dataset PANIC. For open-world semantic segmentation, we use BDDAnomaly~\citep{hendrycks2022icml}, COCO~\citep{lin2014eccv}, SUIM~\citep{islam2020iros}, and PANIC. BDDAnomaly samples from BDD100K and removes all images containing bicycles, motorcycles, and trains from the training set, to create an open-world test set. Similarly, Mask2Anomaly~\citep{rai2024tpami} proposes three open-world splits of COCO, in which $5 \%$, $10 \%$, and $20 \%$ of ground truth categories are discarded from the training set to create an open-world test set, respectively. Classes are removed cumulatively. For the $5 \%$ they remove ``car'', ``cow'', ``pizza'', ``toilet''. Additionally, for the $10 \%$ ``boat'', ``tie'', ``zebra'', ``stop sign''. Finally, for the $20 \%$ ``dining table'', ``banana'', ``bicycle'', ``cake'', ``sink'', ``cat'', ``keyboard'', and ``bear''. Similarly, for the underwater dataset SUIM, we discarded all images containing annotations of ``humans'', ``robots'', and ``wrecks \& ruins''. For open-set panoptic segmentation, we use COCO and PANIC. Finally, for open-world panoptic segmentation, we use PANIC. 

We do not compare our results with open-vocabulary approaches, as we do not employ any large vocabulary, which would require anticipating all categories that could be encountered at test time. The only knowledge we provide to our CNN lies in the limited number of categories that appear in the (closed-world) training set.

\begin{table}
    \small
    \centering
    \makebox[\columnwidth]{%
       \begin{tabular}{cccc} 
        \toprule
        \textbf{Dataset} & \textbf{Modality} & \textbf{mIoU} & \textbf{PQ} \\
        \midrule
        \multirow{2}{*}{Cityscapes} & CW & 71.1 & - \\
        & OW & 68.3 & - \\ 
        \midrule
        \multirow{4}{*}{COCO} & CW & 71.9 & 43.8 \\
        & OW (K $= 5\%$) & 70.0 & 43.6 \\
        & OW (K $= 10\%$) & 68.5 & 39.8 \\
        & OW (K $= 20\%$) & 65.6 & 39.1 \\
        \midrule
        \multirow{2}{*}{BDDAnomaly} & CW & 64.6 & - \\
        & OW & 62.4 & - \\
        \midrule
        \multirow{2}{*}{SUIM} & CW & 68.6 & - \\
        & OW & 60.0 & - \\
      \bottomrule    
    \end{tabular}
     }
     \caption{Results comparison between our model trained in a closed- and open-world fashion on the different datasets we use. Metrics are computed only on known classes.}
     \label{tab:cw_vs_ow}
  \end{table} 
  
  \begin{table}
    \centering
       \begin{tabular}{cccccc} 
        \toprule
        \multirow{3}{*}{\textbf{Approach}} & \multicolumn{2}{c}{\textbf{Pixel-Level}} & \multicolumn{3}{c}{\textbf{Component-Level}} \\
        \cmidrule(lr){2-3}\cmidrule(lr){4-6}
        & AUPR & FPR95 & sIoU & PPV & mF1 \\
        \midrule
        Maskomaly & 93.4 & 6.9 & 55.4 & 51.2 & 49.9 \\
        RbA & 86.1 & 15.9 & 56.3 & 41.4 & 42.0 \\
        ContMAV & 90.2 & 3.8 & 54.5 & 61.9 & 63.6 \\
        UNO & \textbf{96.1} & \textbf{2.3} & \textbf{68.0} & 51.9 & 58.9 \\
        Con2MAV & 90.0 & 2.7 & 59.1 & \textbf{68.3} & \textbf{69.4} \\
      \bottomrule    
    \end{tabular}
     \caption{Anomaly segmentation results on the test set of SegmentMeIfYouCan. Best results are highlighted in bold. More results available on the public leaderboard.}
     \label{tab:anseg_smiyc}
     \vspace{-2em}
  \end{table}

\textbf{Training details and parameters}. In all experiments, we use the one-cycle learning rate policy~\citep{smith2019aimlmdoa} starting from 0.004. We perform random scale, crop, and flip data augmentations, and optimize with Adam~\citep{kingma2015iclr} for 200 epochs with batch size 16. We set loss weights $w_1 = 0.8$, $w_2 = 0.2$, $w_3 = 0.5$, $w_4 = 0.5$, $w_5 = 0.4$, $w_6 = 0.2$, $w_7 = 0.1$, $w_8 = 0.2$, and $w_9 = 0.1$, to have losses in the same order of magnitude. We set $\tau = 0.1$, following \cite{chen2020icml}. For SegmentMeIfYouCan and PANIC, we train on Cityscapes. For BDDAnomaly, COCO, and SUIM, we train on their own training sets.

\subsection{Closed-World Performance}
While designed for open-world segmentation, our approach maintains strong performance on closed-world categories. As shown in~\tabref{tab:cw_vs_ow}, the mIoU and PQ scores on known classes remain comparable when comparing the full open-world model (OW) with its closed-world variant~(CW), where open-world components are removed. Notice that, in all cases, the amount of training data is extremely larger for the closed-world variant with respect to the open-world counterpart. This is due to the fact that, in order to make a dataset suitable for open-world evaluation, images containing certain categories have to be removed from the training set. We report the number of images each training set has in~\tabref{tab:cw_vs_ow}.

Furthermore, as previously discussed, our contributions do not lie in the chosen architecture for the open-world segmentation task, but rather in the loss functions and in the components we use. For the sake of both, training and inference speed, we selected a lightweight CNN for our purposes. However, our contributions can be applied to bigger architectures as well, including vision transformers-based backbones, with no loss of generalizability. Our experiments show that even a smaller, less powerful, fully-convolutional architecture can achieve compelling results, surpassing the performance achieved by more modern architecture on open-world segmentation tasks. 

\begin{table}
    \centering
       \begin{tabular}{cccccc} 
        \toprule
        \multirow{3}{*}{\textbf{Approach}} & \multicolumn{2}{c}{\textbf{Pixel-Level}} & \multicolumn{3}{c}{\textbf{Component-Level}} \\
        \cmidrule(lr){2-3}\cmidrule(lr){4-6}
        & AUPR & FPR95 & sIoU & PPV & mF1 \\
        \midrule
        ContMAV & 91.7 & 66.4 & 15.0 & \textbf{72.1} & 24.2 \\
        Con2MAV & \textbf{95.7} & \textbf{35.3} & \textbf{20.9} & 64.7 & \textbf{31.2} \\
      \bottomrule    
    \end{tabular}
\caption{
Anomaly segmentation results on the hidden test set of our dataset, PANIC. 
Best results are highlighted in bold. Public competition at 
\href{https://www.codabench.org/competitions/4561}{codabench.org/competitions/4561}.
}
     \label{tab:anseg_ours}
  \end{table}

\begin{table}
  \small
  \centering
  \makebox[\columnwidth]{%
  \begin{tabular}{ccc} 
    \toprule
    \textbf{Approach} & AUPR & FPR95 \\
    \midrule
    MaxSoftmax & 3.7 & 24.5  \\
    MC Dropout & 4.3 & 16.6 \\
    Confidence & 3.9 & 24.5 \\
    MaxLogit & 5.4 & 14.0 \\
    ContMAV & 96.1 & 6.9 \\
    \midrule 
    Con2MAV & \textbf{97.9} & \textbf{5.8} \\
    \bottomrule
  \end{tabular}
  }
  \caption{Anomaly segmentation results on BDDAnomaly. Best results are highlighted in bold.}
  \label{tab:anseg_bdd}
\end{table} 

\begin{table}[t]
    \small
    \centering
    \makebox[\columnwidth]{%
    \begin{tabular}{ccc} 
      \toprule
      \textbf{K} $\%$ & \textbf{Approach} & \textbf{mIoU}$_u$ \\
      \midrule
      \multirow{4}{*}{\textbf{5}} 
      & Background + cluster & 8.4\\
      & ContMAV (no feat loss)   & 23.5 \\
      & ContMAV (with feat loss) & 40.0 \\
      & Con2MAV & \textbf{50.5} \\
      \midrule
      \multirow{4}{*}{\textbf{10}} 
      & Background + cluster      & 6.1 \\
      & ContMAV (no feat loss) & 18.8 \\
      & ContMAV (with feat loss) & 38.5\\
      & Con2MAV & \textbf{48.5} \\
      \midrule
      \multirow{4}{*}{\textbf{20}} 
      & Background + cluster & 1.2 \\
      & ContMAV (no feat loss) & 15.0 \\
      & ContMAV (with feat loss) & 33.4 \\
      & Con2MAV & \textbf{48.5} \\
      \bottomrule
    \end{tabular}
    }
    \caption{Open-world semantic segmentation results on the COCO validation set on three different known-unknown splits. K denotes the percentage of unknown classes present in the dataset. Best results are highlighted in bold.}
    \label{tab:owss_coco}
  \end{table}

\subsection{Anomaly Segmentation}
This set of experiments shows that our approach achieves state-of-the-art results on anomaly segmentation. We report results on SegmentMeIfYouCan in~\tabref{tab:anseg_smiyc}, PANIC in~\tabref{tab:anseg_ours}, and BDDAnomaly in~\tabref{tab:anseg_bdd}. We achieve compelling results on all three datasets, consistently outperforming the baselines. Additionally, we outperform our previous approach ContMAV on all datasets. While gains over ContMAV are moderate on SegmentMeIfYouCan and BDDAnomaly, Con2MAV shows a clear advantage on PANIC. Notably, on SegmentMeIfYouCan we rank top 1 in two component-level metrics, outperforming even methods trained with out-of-distribution data.

The performance of ContMAV and Con2MAV on PANIC highlights the complexity and diversity of the benchmark. Compared to SegmentMeIfYouCan, results drop notably, although both methods still achieve good AUPR and PPV. Intuitively, AUPR is high when there is a sharp difference in the probability heatmap between unknown and known areas, while PPV is high when there are few false positive predictions. These results indicate that both approaches behave conservatively on PANIC, producing highly confident anomaly predictions with relatively few false positives, at the cost of missing some unknown objects. We attribute this behavior to the large variety of unknown categories and object instances represented in PANIC compared to existing anomaly-segmentation benchmarks.

\begin{table}[t]
    \small
    \centering
    \resizebox{1\linewidth}{!}{
    \begin{tabular}{ccccc} 
      \toprule
      \multirow{3}{*}{\textbf{Approach}} & \multicolumn{3}{c}{\textbf{IoU}$_u$} & \multirow{3}{*}{\textbf{mIoU}$_u$}\\
      \cmidrule(lr){2-4}
      & Train & Motorcycle & Bicycle & \\
      \midrule
      Background + cluster      & 0 & 32.3 & 32.8 & 21.7\\
      ContMAV & 62.4 & 62.2 & \textbf{56.8} & 60.5 \\
      Con2MAV & \textbf{66.5} & \textbf{64.4} & 53.8 & \textbf{61.6}\\
      \midrule 
      Closed-world  & 72.3 & 69.3 & 60.9 & 67.5\\
      \bottomrule
    \end{tabular}
    }
    \caption{Open-world semantic segmentation results on BDDAnomaly. Best results are highlighted in bold.}
   \label{tab:owss_bdd}
  \end{table}

\begin{table}[t]
  \small
  \centering
  \resizebox{1\linewidth}{!}{
  \begin{tabular}{ccccc} 
    \toprule
    \multirow{3}{*}{\textbf{Approach}} & \multicolumn{3}{c}{\textbf{IoU}$_u$} & \multirow{3}{*}{\textbf{mIoU}$_u$}\\
    \cmidrule(lr){2-4}
    & Human & Wrecks \& Ruins & Robot \\
    \midrule
    ContMAV & 46.2 & 36.9 & 46.2 & 43.1 \\
    Con2MAV & \textbf{69.4} & \textbf{64.3} & \textbf{53.0} & \textbf{62.2} \\
    \midrule 
    Closed-world  & 82.0 & 71.6 & 80.2 & 77.9 \\
    \bottomrule
  \end{tabular}
  }
  \caption{Open-world semantic segmentation results on the SUIM dataset. Best results are highlighted in bold.}
 \label{tab:owss_suim}
\end{table}

\begin{table}[t]
    \small
    \centering
    \makebox[\columnwidth]{%
    \begin{tabular}{cccc} 
      \toprule
      \multirow{3}{*}{\textbf{Approach}} & \multicolumn{3}{c}{\textbf{Unknown Classes}} \\
      \cmidrule(lr){2-4}
      &  mIoU$_u$ & Com & Hom \\
      \midrule
      ContMAV & 15.8 & \textbf{81.4} & 77.3 \\
      Con2MAV & \textbf{20.2} & 77.6 & \textbf{78.3} \\
      \bottomrule
    \end{tabular}
    }
    \caption{Open-world semantic segmentation results on the hidden test set of our dataset, PANIC. ``Com'' and ``Hom'' stand for completeness and homogeneity, respectively. Best results are highlighted in bold. Public competition is available at \href{https://www.codabench.org/competitions/4563}{codabench.org/competitions/4563}.}
    \vspace{-1.5em}
   \label{tab:owss_ours}
  \end{table}

\begin{figure*}[t!]
    \centering
    \includegraphics[width=0.9\linewidth]{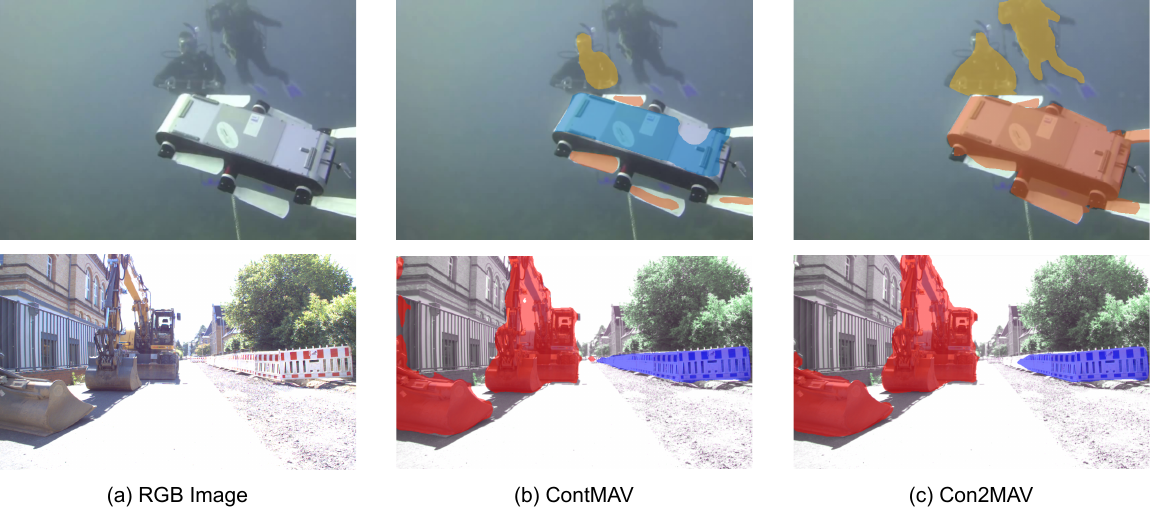}
    \caption{Qualitative results of our approach, Con2MAV, on open-world semantic segmentation on SUIM (top row) and PANIC (bottom row). The prediction mask is overlayed to the input RGB for clarity. In the prediction, different colors correspond to different predicted classes. We compare our approach, Con2MAV (right), with our old method, ContMAV (center). \vspace{-1em}}
    \label{fig:exp_ss}
\end{figure*}

\subsection{Open-World Semantic Segmentation}
The second set of experiments is about open-world semantic segmentation. The task is uncommon in literature, so we compare against ContMAV~\citep{sodano2024cvpr}, our previous approach designed specifically for open-world semantic segmentation. We include a lower-bound baseline that treats unknowns as background and applies K-means clustering in the background feature space to identify unknown classes. For this baseline, the number of clusters is set to the ground-truth number of unknown categories, providing an approximate lower bound on feature-space clustering performance under favorable conditions. We report results on COCO in~\tabref{tab:owss_coco}, BDDAnomaly in~\tabref{tab:owss_bdd}, SUIM in~\tabref{tab:owss_suim}, and PANIC in~\tabref{tab:owss_ours}. On COCO, we consistently outperform ContMAV by at least $10 \%$ in the mIoU computation. On BDDAnomaly, we outperform ContMAV in two out of three unknown classes, while getting close to the performance of the closed-world model, reported as a performance upper bound. On PANIC, we improve ContMAV by almost $5 \%$ mIoU, despite our previous approach achieves a better completeness. 

The use of the SUIM dataset is to test how our approach performs when only few known training classes are available. In fact, previous approaches like ContMAV~\citep{sodano2024cvpr} suffer from this issue, as their category descriptors are not powerful enough. In SUIM, we reduce the number of training classes to just $4$. The results in~\tabref{tab:owss_suim} show that our approach succesfully addressed this issue of having few available training categories, as we outperform the baseline by $19 \%$ mIoU, and up to $28 \%$ in a single individual category (``wreck \& ruins''). We show qualitative results for this task in~\figref{fig:exp_ss}.

\begin{table}[t]
    \small
    \centering
    \makebox[\columnwidth]{%
       \begin{tabular}{ccccc} 
        \toprule
        \multirow{3}{*}{\textbf{K $\%$}} & \multirow{3}{*}{\textbf{Approach}} & \multicolumn{3}{c}{\textbf{Unknown Classes}} \\
        \cmidrule(lr){3-5}
        && $\mathrm{PQ}_{u}$ & $\mathrm{SQ}_{u}$ & $\mathrm{RQ}_{u}$ \\ 
        \midrule
        \multirow{4}{*}{\textbf{5}} & Void-train & 8.6 & 72.7 & 11.8 \\
        & EOPSN & 23.1 & 74.7 & 30.9 \\
        & Mask2Anomaly & 24.3 & 78.2 & 32.1 \\
        & Con2MAV  & \textbf{25.1} & \textbf{78.4} & \textbf{34.6} \\ 
        \midrule
        \multirow{4}{*}{\textbf{10}} & Void-train & 8.1 & 72.6 & 11.2 \\
        & EOPSN & 17.9 & 76.8 & 23.3 \\
        & Mask2Anomaly & 19.7 & 77.0 & 25.7 \\
        & Con2MAV  & \textbf{21.3} & \textbf{77.1} & \textbf{27.6} \\ 
        \midrule
        \multirow{4}{*}{\textbf{20}} & Void-train & 7.5 & 72.9 & 10.3 \\
        & EOPSN & 11.3 & 73.8 & 15.3 \\
        & Mask2Anomaly & 14.6 & 76.2 & 19.1 \\
        & Con2MAV  & \textbf{22.1} &  \textbf{88.5} & \textbf{24.9} \\ 
      \bottomrule    
    \end{tabular}
     }
     \caption{Open-set panoptic segmentation results on the COCO validation set on three different known-unknown splits. K denotes the percentage of unknown classes present in the dataset. Best results are highlighted in bold.}
     \label{tab:osps_coco}
  \end{table}

\subsection{Open-Set Panoptic Segmentation}
\begin{figure*}[t!]
    \centering
    \includegraphics[width=0.95\linewidth]{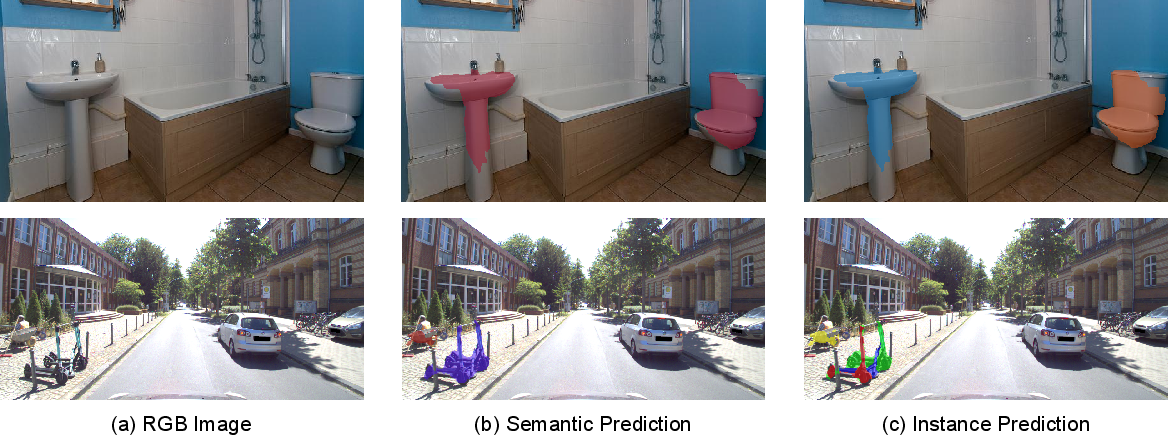}
    \caption{Qualitative results of our approach, Con2MAV, on open-set panoptic segmentation on COCO (top row), and PANIC (bottom row). The prediction mask is overlayed to the input RGB for clarity. In the semantic prediction, the colored area indicates the anomalous region. In the instance prediction, different colors correspond to different instance ids.}
    \label{fig:exp_osps}
\end{figure*}

Here, we present experiments on open-set panoptic segmentation. For this task, we use COCO and PANIC. On COCO, we compare with other state-of-the-art approaches and report results in~\tabref{tab:osps_coco}. We consistently outperform all baselines in all metrics. Additionally, while the baselines drop in performance when increasing the open-world set dimension (i.e., when $K$ grows), our performance remains similar. We report results on PANIC in~\tabref{tab:osps_ours}, where we compare with ContMAV~\citep{sodano2024cvpr} augmented with the same instance decoder introduced in~\secref{subsec:approach}. It is interesting to note that our performance on PANIC is quite similar to what we achieve on COCO, especially in the more challenging splits ($10\%$ and $20 \%$), that are also characterized by a large number of unknown objects. We note that concurrent work, Prior2Former~\citep{schmidt2025iccv}, published after the first submission of this paper, has recently reported stronger results on the PANIC leaderboard, highlighting both the rapid progress in open-set panoptic segmentation and the relevance of PANIC as a benchmark for evaluating such methods. We show qualitative results for this task in~\figref{fig:exp_osps}.

\begin{table}[t]
    \centering
    \makebox[\columnwidth]{%
       \begin{tabular}{cccc} 
        \toprule
        \multirow{3}{*}{\textbf{Approach}} & \multicolumn{3}{c}{\textbf{Unknown Classes}} \\
        \cmidrule(lr){2-4}
        & $\mathrm{PQ}_{u}$ & $\mathrm{SQ}_{u}$ & $\mathrm{RQ}_{u}$ \\ 
        \midrule
        ContMAV + ins. seg. & 18.4 & 71.7 & 24.7 \\
        Prior2Former & \textbf{52.9} & \textbf{87.1} & \textbf{56.7} \\
        Con2MAV (ours) & 21.6 & 72.4 & 28.4 \\ 
      \bottomrule    
    \end{tabular}
    }
    \caption{Open-set panoptic segmentation results on the hidden test set of our dataset, PANIC. Public competition is available at \href{https://www.codabench.org/competitions/4562}{codabench.org/competitions/4562}.}
    \label{tab:osps_ours}
\end{table}

\begin{table}[t]
    \centering
    \small
    \makebox[\columnwidth]{%
    \begin{tabular}{ccccc} 
      \toprule
      \multirow{3}{*}{\textbf{Approach}} & \multicolumn{4}{c}{\textbf{Unknown Classes}} \\
      \cmidrule(lr){2-5}
      & $\mathrm{PQ}_u$ & $\mathrm{mIoU}_{u}$ & $\mathrm{Com}$ & $\mathrm{Hom}$  \\ 
      \midrule
      ContMAV + ins. seg. & 21.3 & 15.8 & 75.4 & 77.3 \\
      Con2MAV (ours)  & \textbf{25.1} & \textbf{20.2} & \textbf{77.6} & \textbf{78.3} \\ 
      \bottomrule    
    \end{tabular}
    }
    \caption{Open-world panoptic segmentation results on the hidden test set of our dataset, PANIC. Public
    competition is available at \href{https://www.codabench.org/competitions/4477}{codabench.org/competitions/4477}.}
    \label{tab:owps_ours}
  \end{table}

\begin{figure*}[t!]
    \centering
    \includegraphics[width=0.95\linewidth]{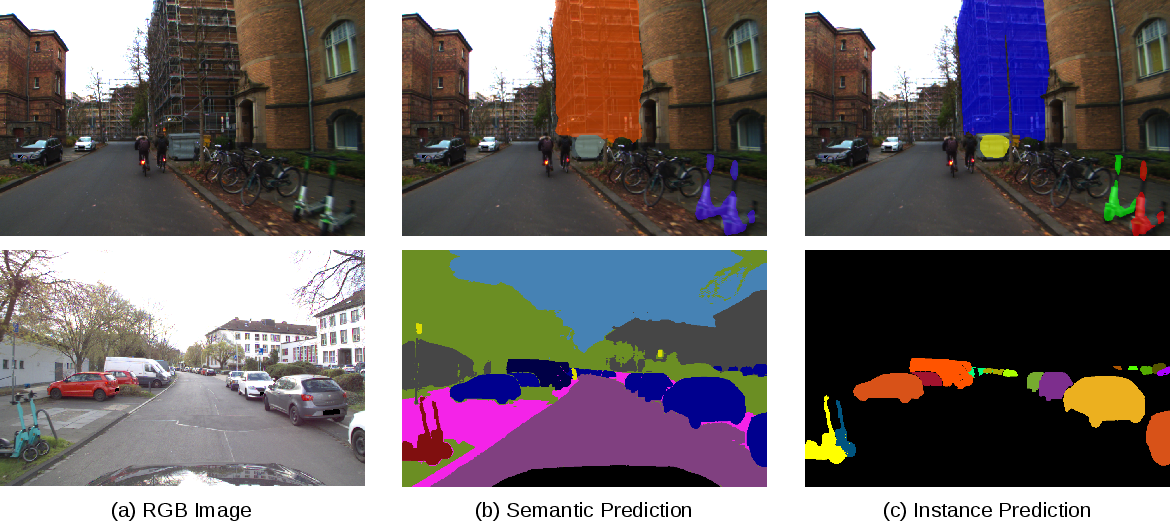}
    \caption{Qualitative results of our approach, Con2MAV, on open-world panoptic segmentation on PANIC. The prediction mask is overlayed to the input RGB for clarity. In the semantic prediction, different colors correspond to different predicted classes. In the instance prediction, different colors correspond to different instance ids. In the top row, we show the segmentation of the unknown parts only. In the bottom row, the complete open-world panoptic segmentation results.}
    \label{fig:exp_owps}
  \end{figure*}

\subsection{Open-World Panoptic Segmentation}
The last experiment shows our performance on open-world panoptic segmentation on PANIC. The results are reported in~\tabref{tab:owps_ours}. For this novel task, there is no baseline to compare to. As in the previous case, we augment ContMAV~\citep{sodano2024cvpr} with the same instance decoder we introduced in~\secref{subsec:approach}. Our approach achieves $24.3 \%$ panoptic quality, outperforming the baseline, and especially has good performance when it comes to completeness and homogeneity. This means that the categories we predict do not span over multiple ground truth classes but stay close to a single one, and that our ground truth categories are not predicted as many categories at once, but the prediction is quite consistent. We show qualitative results in~\figref{fig:exp_owps}.

\begin{table}[t]
    \makebox[\columnwidth]{%
  \begin{tabular}{ccccc}
    \toprule
    & \multirow{3}{*}{$\b{D}_C$} & \multirow{3}{*}{\textbf{PL}} & \multicolumn{2}{c}{\textbf{Unknown Classes}} \\
    \cmidrule(lr){4-5}
    & & & $\mathrm{mIoU}_u$ & $\mathrm{PQ}_u$ \\
    \midrule
    A &  &  &  42.3 & 8.3 \\
    B &  & \cmark & 48.2 & 11.9 \\
    C & \cmark&&  50.2 & 15.8 \\
    \midrule
    D & \cmark & \cmark & 57.3 & 22.4 \\
    \bottomrule
  \end{tabular}
    }
  \caption{Ablation study on decoder components. Last row is our approach, Con2MAV. \vspace{-1em}}
  \label{tab:ablation_1}
\end{table}

\subsection{Ablation Studies}
Finally, we provide ablation studies to investigate the individual contribution of the modules we introduced. We perform all ablation studies on BDDAnomaly since we have access to test set ground truth labels.

\begin{table}[t]
  \centering
  \small
  \makebox[\columnwidth]{%
  \begin{tabular}{cccccc}
    \toprule
    &\multirow{3}{*}{\textbf{Offset}} & \multirow{3}{*}{\textbf{Divergence}} & \multirow{3}{*}{\textbf{Curl}} & \multicolumn{2}{c}{\textbf{Unknown Classes}} \\
    \cmidrule(lr){5-6}
    && & & $\mathrm{mIoU}_u$ & $\mathrm{PQ}_u$ \\
    \midrule
    E & \cmark & & & 57.1 & 12.4 \\
    F & \cmark & \cmark &  & 56.7 & 17.8 \\
    G & \cmark & & \cmark & 56.9 & 15.0 \\
    \midrule
    H & \cmark & \cmark & \cmark & 57.3 & 22.4 \\
    \bottomrule
  \end{tabular}
  }
  \caption{Ablation study on loss functions for segmenting unknown instances. Last row is our approach, Con2MAV. \vspace{-1em}}
  \label{tab:ablation_2}
\end{table}

\begin{table}[t]
    \makebox[\columnwidth]{%
  \begin{tabular}{ccc} 
    \toprule
    \textbf{Approach} & GFLOPs & Params \\
    \midrule
    Maskomaly & 937 & 215M  \\
    Mask2Anomaly & 258 & \textbf{23M} \\
    ContMAV & 84 & 48M \\
    \midrule
    Con2MAV & \textbf{50} & 65M \\
    \bottomrule
  \end{tabular}
    }
  \caption{Details for architectural efficiency measured in number of parameters and GFLOPs. \vspace{-2em}}
  \label{tab:efficiency}
\end{table}

\begin{figure*}[ht!]
    \centering
    \includegraphics[width=0.9\textwidth]{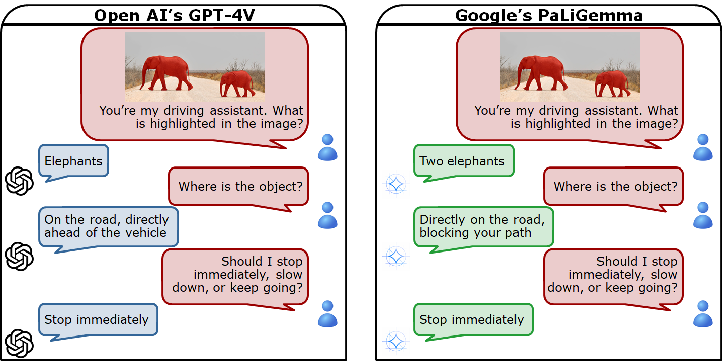}
    \caption{GPT-4V and PaliGemma results on an image from SegmentMeIfYouCan~\citep{chan2021neurips}.}
    \label{fig:fm1}
    
    \vspace{1em} 
    
    \includegraphics[width=0.9\textwidth]{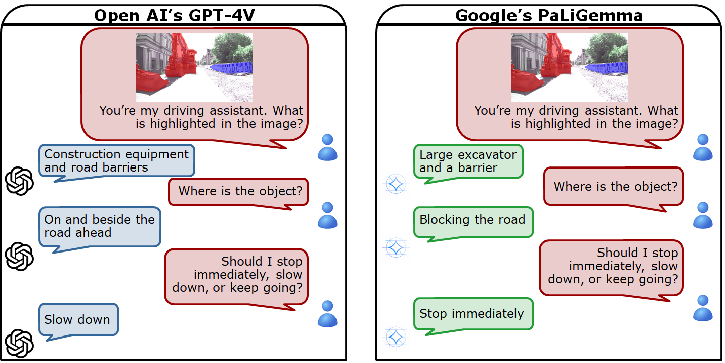}
    \caption{GPT-4V and PaliGemma results on an image from PANIC (ours), showing the interaction between the user and the respective visual-language model. Answers and interactions are unabbreviated, and shown as chats for illustrative purposes.}
    \label{fig:fm2}
\end{figure*}

The first ablation study shows the contribution of (i) the contrastive decoder (called $\b{D}_C$ in the table), and (ii) the pre-logit instead of the pre-softmax features for computing the loss functions of the semantic and contrastive decoder (called PL). The contrastive decoder plays a key role in achieving strong performance by complementing the semantic decoder and reinforcing its predictions (see entries C and D vs. A and B in the table). It acts as a “stabilizer” for open-world semantic segmentation, benefiting from its simpler objective, anomaly segmentation, which makes it more robust to noise and errors. Applying the loss at the pre-logit level, rather than on the low-dimensional pre-softmax features, also yields clear benefits. This design choice improves performance by more than $5\%$ on both metrics, both without the contrastive decoder (B vs. A) and with it (D vs. C).

The second ablation study evaluates the impact of our vector field-inspired loss function for the instance segmentation. We compare our full model with one lacking both divergence and curl (H in the table), or using only one (I and J). Results are shown in~\tabref{tab:ablation_2}. The mIoU is always very similar, because these loss functions do not have any impact on the semantic segmentation. The small differences are due to the fact that they all affect the shared encoder through backpropagation. Both divergence and curl individually improve the basic model. Additionally, the experiment suggests that divergence is more useful than curl for instance segmentation. Combining the two proves to be the best possible combination (K).

\subsection{Architectural Efficiency}
As pointed out in~\secref{subsec:network_arch}, we designed our neural network in order to be lightweight. In~\tabref{tab:efficiency}, we report the number of parameters and the GFLOPs of our model together with three state-of-the-art models with code available from the SegmentMeIfYouCan public benchmark. We show that our architecture is competitive and performs well in terms of efficiency.

\section{What Could This Enable?} \label{sec:vlms}
As already mentioned, open-world panoptic segmentation shares similarities with open-vocabulary segmentation, that aims to segment novel classes and objects by leveraging large language models (LLMs) for class discovery. However, open-vocabulary approaches typically require explicit category prompts to guide the segmentation process. When we evaluated such methods~\citep{zou2023cvpr} on PANIC, performance was poor unless the target categories were clearly specified. This, however, defeats the purpose of identifying unknown objects autonomously, without any prior knowledge. 

To address this limitation, we used a complementary strategy. We fed our open-world panoptic segmentation predictions to two vision-language models (VLMs): \mbox{GPT-4V}~\citep{openai2023} and PaliGemma~\citep{steiner2024arxiv}. We provided the input RGB image overlayed with our open-world prediction, and queried the VLMs to identify the unknown objects and suggest appropriate actions for the driving scenario. We show two examples (\figref{fig:fm1} and~\figref{fig:fm2}), organized as chat interactions for visualization.
These results highlight an intriguing direction: models like ours, capable of detecting unknown instances without prior class information, can serve as a front-end to VLMs for visual prompting, enabling contextual understanding and decision-making in complex, open-world environments.

\section{Conclusion}\label{sec:conclusion}
In this work, we introduce Con2MAV, an approach for open-world panoptic segmentation, and PANIC, a new benchmark for several open-world segmentation tasks in the autonomous driving domain. Through extensive quantitative experiments on multiple datasets, we demonstrate the effectiveness of our Con2MAV approach across different domains, such as autonomous driving and underwater monitoring, on all four tasks that we introduce. Furthermore, our experiments on PANIC demonstrate its suitability as a comprehensive benchmark, covering a large variety of unknown categories and object instances for evaluating open-world perception methods. We believe this work, and especially the creation of a dedicated benchmark for all kinds of open-world segmentation tasks, will facilitate the development of novel approaches and encourage further advancements in the field.

\begin{funding}
This work has partially been funded 
  by the Deutsche Forschungsgemeinschaft (DFG, German Research Foundation) under Germany's Excellence Strategy, EXC-2070 -- 390732324 (PhenoRob), and by the German Federal Ministry of Education and Research (BMBF) in the project ``Robotics Institute Germany'', grant No.~16ME0999.
\end{funding}

\bibliographystyle{SageH}
\bibliography{references}

\end{document}